\documentclass[lettersize,journal]{IEEEtran}

\usepackage[caption=false,font=normalsize,labelfont=sf,textfont=sf]{subfig}

\usepackage{textcomp}
\usepackage{stfloats}
\usepackage{url}
\usepackage{verbatim}
\usepackage{graphicx}
\usepackage{cite}

\usepackage[utf8]{inputenc} 
\usepackage[T1]{fontenc}    
\usepackage{hyperref}       
\usepackage{url}            
\usepackage{booktabs}       
\usepackage{amsfonts}       
\usepackage{nicefrac}       
\usepackage{microtype}      
\usepackage{xcolor}         
\usepackage{amsmath}
\newcommand{\xmark}{\ding{55}}%
\newcommand{\noo}{\textcolor{red}{\xmark}}
\newcommand{\yes}{\textcolor{OliveGreen}{\checkmark}}
\usepackage{colortbl} 
\usepackage{arydshln}
\usepackage[dvipsnames]{xcolor}
\usepackage{pifont} 
\usepackage{tikz} 
\definecolor{tabfirst}{rgb}{1, 0.75, 0.7}
\definecolor{tabsecond}{rgb}{1, 0.83, 0.7}
\definecolor{tabthird}{rgb}{1, 0.96, 0.7}
\usepackage{multirow}
\usepackage{multicol}
\usepackage{subcaption}

\title{VPGS-SLAM: Voxel-based Progressive 3D Gaussian SLAM in Large-Scale Scenes}

\author{%
  Tianchen Deng, Wenhua Wu,  Junjie He, Yue Pan, Shenghai Yuan,\\  Danwei Wang,~\IEEEmembership{Life Fellow,~IEEE}, Hesheng Wang~\IEEEmembership{Senior Member,~IEEE}
\thanks{Tianchen Deng, Wenhua Wu, Hesheng Wang are with the School of Automation and Intelligent Sensing, Shanghai Jiao Tong University, and Key Laboratory of
System Control and Information Processing, Ministry of Education, Shanghai 200240, China. Junjie He is at the Thrust of Robotics and Autonomous Systems, The Hong Kong University of Science and Technology (Guangzhou). Yue Pan is with the University of Bonn. Danwei Wang and Shenghai Yuan are with the School of Electrical and Electronic Engineering, Nanyang Technological University, Singapore. This research is supported by the National Research Foundation, Singapore, under the NRF Medium Sized Centre scheme (CARTIN), Maritime and Port Authority of Singapore under its Maritime Transformation Programme (Project No. SMI-2022-MTP-04), ASTAR under National Robotics Programme with Grant No.M22NBK0109. The first two authors contibute equal to this paper.  (*corresponding author: wanghesheng@sjtu.edu.cn)
	}
}
\begin{document}

\maketitle

\begin{abstract}
  3D Gaussian Splatting has recently shown promising results in dense visual SLAM. However, existing 3DGS-based SLAM methods are all constrained to small-room scenarios and struggle with memory explosion in large-scale urban scenes and long sequences. To this end, we propose VPGS-SLAM, a novel 3DGS-based large-scale RGBD SLAM framework for both indoor and outdoor scenarios. We design a novel voxel-based progressive 3D Gaussian mapping method with multiple submaps for  compact and accurate scene representation in large-scale and long-sequence scenes. This allows us to scale up to arbitrary scenes and improves robustness (even under pose drifts). In addition, we propose a 2D-3D fusion camera tracking method to achieve robust and accurate camera tracking in both indoor and outdoor large-scale scenes. Furthermore, we design a 2D-3D Gaussian loop closure method to eliminate pose drift. We further propose a submap fusion method with online distillation to achieve global consistency in large-scale scenes when detecting a loop. Experiments on various indoor and outdoor datasets demonstrate the superiority and generalizability of the proposed framework.
  The code will be open-sourced on \href{https://github.com/dtc111111/vpgs-slam}{https://github.com/dtc111111/vpgs-slam}.

\end{abstract}

 \begin{figure*}[ht]
    \centering
    
    \includegraphics[width=\linewidth]{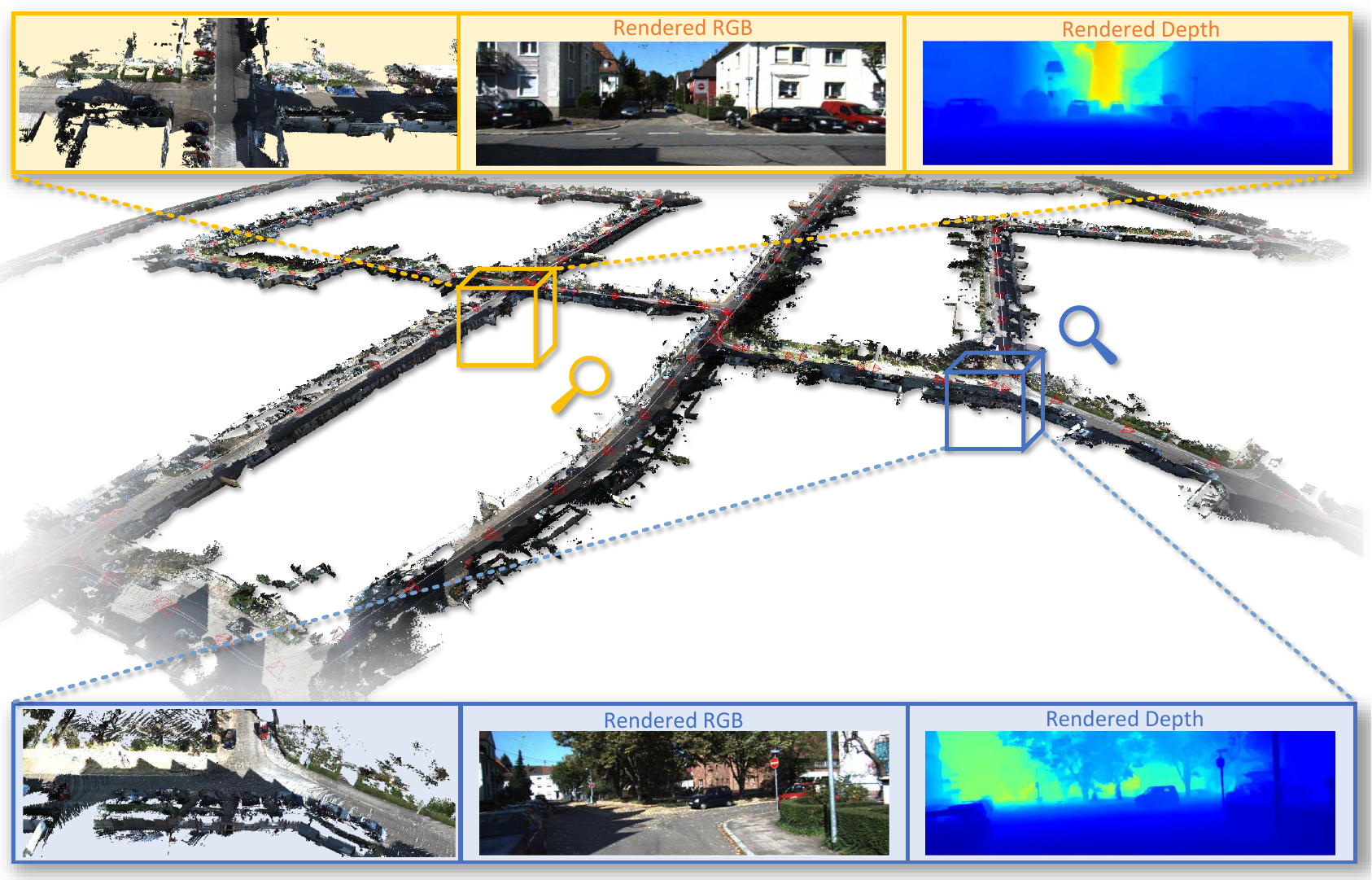}
    
    \caption{We present VPGS-SLAM, a novel large-scale SLAM framework with voxel-based progressive 3D Gaussian representation, 2D-3D assisted camera tracking and 3D Gaussian loop closure.
  Depicted in the middle, we demonstrate the large-scale globally consistent 3D Gaussian map built with our approach. At the top and bottom of the figure, we include zoomed-in views of the map with RGB and depth images rendered by our method, indicated by dashed blue and yellow boxes.}

 \label{fig:teaser}
\end{figure*}


\section{Introduction}

\label{sec:intro}
Visual Simultaneous localization and mapping (SLAM)~\cite{past,10648881} 
has been a fundamental problem with wide applications such as autonomous driving~\cite{tits1}, robotics~\cite{deng2025best3dscenerepresentation}, and remote sensing.  

In intelligent transportation systems, accurate pose estimation constitutes a fundamental prerequisite for safe and reliable autonomous driving. Errors in vehicle pose estimation can directly compromise motion planning and control, leading to hazardous behaviors such as premature or unnecessary braking, unsafe following distances, and incorrect obstacle avoidance decisions. More critically, the consequences of pose misestimation extend beyond the affected vehicle itself. In real-world traffic environments, abnormal driving actions triggered by inaccurate pose estimates can influence surrounding vehicles through interaction and response, potentially propagating as cascading unsafe maneuvers, large-scale braking waves, and traffic flow instability. Such error amplification mechanisms may significantly elevate the risk of traffic accidents and degrade overall transportation safety, underscoring the essential role of robust and accurate pose estimation in autonomous driving systems.

Several traditional methods\cite{orbslam2,vins,xie,cte,VPL-SLAM,lightslam} have been introduced over the years. They use handcraft descriptors for image matching and represent scenes using sparse feature point maps. Due to the sparse nature of such point cloud, it is difficult for humans to understand how machines interact with the scene, and these methods cannot meet the demands of collision avoidance and motion planning. Attention then turns to dense scene reconstruction, exemplified by DTAM~\cite{dtam}, Kintinuous~\cite{Kintinuous}, and ElasticFusion~\cite{elasticfusion}. However, their performance is limited by high memory consumption, slow processing speeds.

Following the introduction of Neural Radiance Fields 
(NeRF), numerous research efforts have focused on combining implicit scene representation with SLAM systems and autonomous driving~\cite{qin2024crowd}. iMAP~\cite{imap} pioneered the use of a single MLP to represent the scene, while \cite{niceslam,eslam,coslam,plgslam} have further enhanced scene representation through hybrid feature grids, axis-aligned feature planes, joint coordinate-parametric encoding, and multiple implicit submaps. NeSLAM~\cite{neslam} use a depth completion and denoising method to improve scene representation.
To further improve rendering speed, recent methods have started to explore 3D Gaussian Splatting (3DGS)~\cite{3dgs} in SLAM systems, as demonstrated by \cite{splatam,monogs}. GS-based SLAM methods leverage a point-based representation associated with 3D Gaussian attributes and adopt the rasterization pipeline to
render the images, achieving fast rendering speed and promising image quality. 

However, existing SLAM systems primarily focus on small-scale indoor environments; they face significant challenges in representing large-scale scenes (e.g., multi-room apartments and urban scenes). Some  We outline the key challenges for indoor and outdoor large-scale 3DGS-based SLAM systems: \textit{\textbf{a) Redundant 3D Gaussian ellipsoids:}} existing methods employ a substantial number of 3D Gaussian ellipsoids to represent the scene. These methods typically require more than 500MB to represent a small room-scale scene, which severely limits their applicability and results in memory overload in large-scale environments. \textit{\textbf{b) Accumulation of errors and pose drift:}} Existing works rely on rendered loss for camera tracking, which proves inaccurate and unreliable in large-scale real-world environments with dramatic movement, motion blur and exposure change. \textit{\textbf{c) Global inconsistency:}} In long sequences and large-scale scenes, when the robot revisits the same location, it is essential to ensure  spatial correlations, and long-term memory for global consistency.

To this end, we propose VPGS-SLAM, a novel 3DGS-based large-scale SLAM framework with voxel-based progressive 3D Gaussian representation, 2D-3D fusion camera tracking, and 2D-3D  Gaussian based loop detection and correction, and submap fusion.
We use a collection of submaps to represent the entire scene, which dynamically initializes local scene representation when the camera moves to the bounds of the local submaps. The entire scene is divided into multiple local submaps, which can significantly improve the scene representation capacity of large-scale scenes. The submap parameters do not need to be retained in memory after optimization, which can significantly reduce online memory requirements and enhances the scalability of the framework.
 \begin{figure*}[ht]
    \centering
    
    \includegraphics[width=\linewidth]{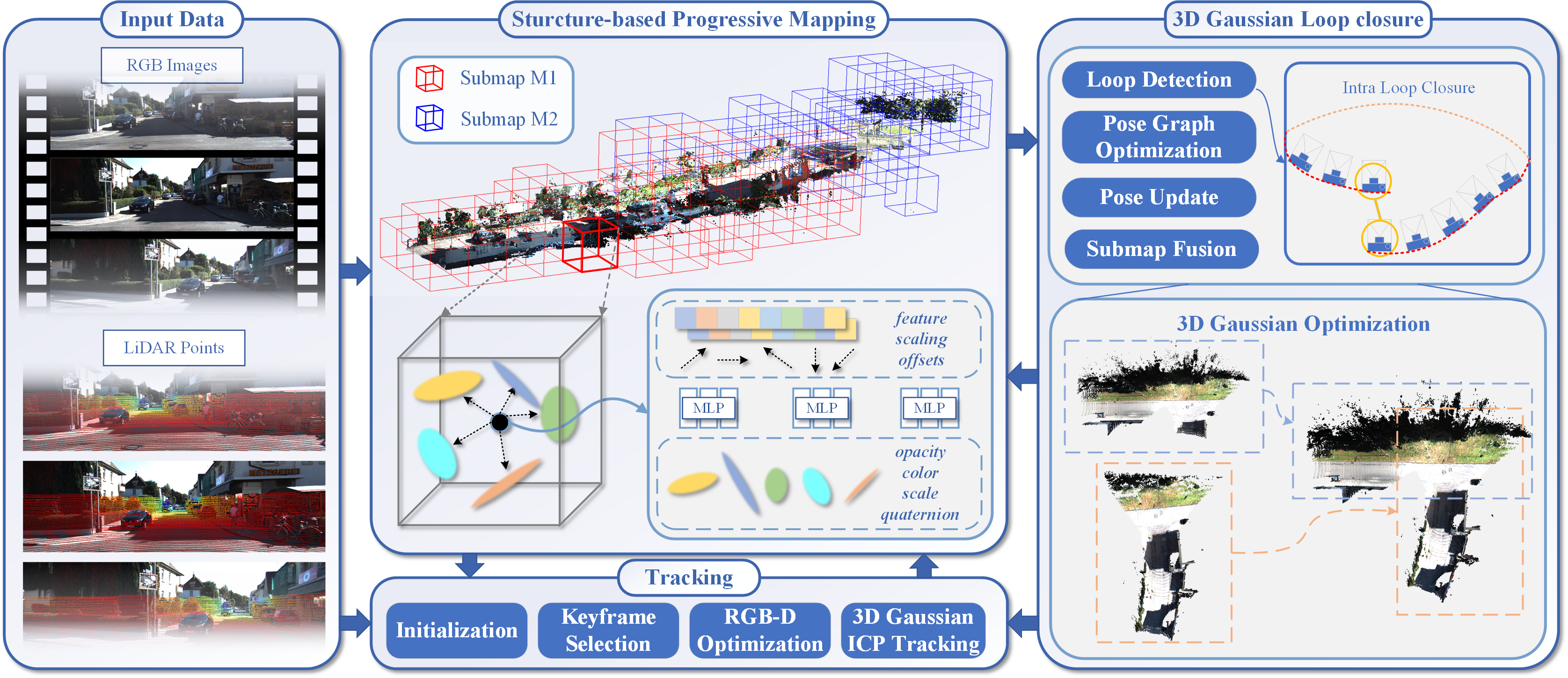}
   
    \caption{\textbf{System Overview.} Our system is a large-scale SLAM framework, with voxel-based progressive 3D Gaussian representation, 2D-3D fusion camera tracking, and 3D Gaussian loop closure. Our framework takes color images and 3D point clouds as input. Our method can achieve accurate and efficient scene reconstruction, camera tracking, and global map generation.}
 
    \label{fig:system}
\end{figure*}

In local scene representation, a sequentialized voxel-based 3D Gaussian representation is proposed tailored for online SLAM framework. We design an efficient hybrid data structure that combines the multi-resolution voxel representation with 3D Gaussian ellipsoids. The local scene is sequentially initialized with sparse voxels. When a keyframe arrives, we initialize voxels in the regions observed by this frame. Each voxel is assigned a corresponding anchor point. Each anchor spawns a set of neural Gaussians with learnable offsets, where attributes such as opacity, color, rotation, and scale are predicted based on the feature of the anchor point and the viewing position. We leverage scene geometry structure to guide and constrain the distribution of 3D Gaussian ellipsoids. We also design a novel sequentially growing and pruning strategy for voxels and anchor points.

For camera tracking, we propose a 2D-3D fusion tracking method to combine the 3D Gaussian ellipsoids geometric information with 2D photometric information. We optimize the pose with 2D photometric loss using RGB and depth loss in the coarse-level optimization stage. We then incorporate a 3D voxel based ICP method to perform frame-to-map pose estimation in the fine-level stage. We provide a good initial estimate for 3D point matching through the 2D rendering loss, enabling further accurate pose estimation. To handle both indoor and outdoor environments, we propose an adaptive 2D information assessing and parameter selection strategy. By assessing the richness of 2D information in the scene, our method will choose different key parameters in the tracking and loop closure modules, enabling robust performance across diverse environments. Moreover, to alleviate the cumulative pose drift in large-scale scenes, we propose a 2D-3D Gaussian based loop closure method with loop detection, pose graph optimization with voxel ICP and rendering loss. We also design an online distillation method for submap fusion that is triggered upon loop closure detection. This method merges the multiple submaps involved in the loop, thereby enhancing global consistency of the global map.
\textbf{Overall, our contributions are shown as follows:}
\begin{itemize}
    \item We propose VPGS-SLAM, a novel 3DGS-based large-scale RGBD SLAM framework that enables efficient and accurate scene reconstruction and pose estimation in both indoor and city-scale environments.
    
    \item A novel sequentialized voxel-based progressive 3D Gaussian representation is proposed for compact and efficient scene representation in large-scale scenes. We design multiple local submaps with multi-resolution voxel representation to achieve efficient and accurate reconstruction in large-scale scenes.
    \item A novel 2D-3D fusion tracking method is proposed to combine the 3D Gaussian geometric information with 2D photometric information for accurate pose estimation. We also propose a 2D-3D Gaussian loop closure method with loop detection, and pose graph optimization.  A submap fusion method with online distillation is proposed to achieve global consistency. Experiments on various indoor and outdoor datasets demonstrate the superiority of the proposed method in both mapping and tracking.
\end{itemize}

\section{Related Work}
\noindent\textbf{Traditional SLAM.} SLAM has been an active research field for the past two decades. Traditional visual SLAM algorithms~\cite{orbslam} estimate accurate camera poses and represent the scene using sparse point clouds. \cite{10786862} utilizes tightly-coupled LiDAR-Visual-Inertial odometry with multi-modal semantic information to enhance the robustness and accuracy of SLAM. DTAM~\cite{dtam} was the first RGB-D approach to achieve dense scene reconstruction. Some learning-based methods~\cite{droidslam,dvlo}, integrate traditional geometric frameworks with deep learning networks for improved camera tracking and mapping.For 3D LiDAR odometry and mapping, similar to feature-matching methods widely used in visual SLAM, the seminal work LOAM~\cite{loam} proposes extracting sparse planar or edge feature points from the scan point cloud and registering them to the previous frame or the feature point map using ICP. Recently, CT-ICP~\cite{cticp} and KISS-ICP~\cite{kissicp} have achieved robust LiDAR odometry performance without the need for feature point extraction. 

\noindent\textbf{NeRF-based SLAM.} With the introduction of NeRF\cite{NeRF}, iMAP\cite{imap} pioneered the use of a single multi-layer perceptron (MLP) to represent the scene, while NICE-SLAM~\cite{niceslam} introduced learnable hierarchical feature grids. ESLAM~\cite{eslam} and Co-SLAM~\cite{coslam} further enhanced scene representation using tri-planes and joint coordinate-parametric encoding. Some methods~\cite{plgslam,incremental,s2kan} proposed a novel progressive scene representation that dynamically allocates new local representations. Go-SLAM~\cite{goslam}, Loopy-SLAM~\cite{loopyslam} use loop closure to enhance the camera tracking performance. SNI-SLAM~\cite{snislam} leverages semantic information. \cite{pointslam,glorieslam,nerfloam,pinslam} use neural point-based neural radiance fields for large-scale scenes and high-accuracy reconstruction. DDN-SLAM~\cite{ddn-slam} focus on dynamic scene representation with masks. Unlike these methods, which use neural implicit features, our approach adopts an explicit 3D Gaussian representation, significantly improving the scalability of our method.

\noindent\textbf{GS-based SLAM.} Recently, 3D Gaussian Splatting ~\cite{3dgs} has emerged using 3D Gaussians as primitives for real-time neural rendering. SplaTAM~\cite{splatam}, MonoGS~\cite{monogs}, Gaussian-SLAM~\cite{gaussianslam}, and other works~\cite{gsslam, compactslam,loopsplat, gs-icp, splatslam,rtgslam,sgsslam,mg-slam,densesplat} are the pioneer works that successfully combine the advantages of 3D Gaussian Splatting with SLAM. These methods achieve fast rendering speed and high-fidelity reconstruction performance. GigaSLAM~\cite{gigaslam} is the concurrent work of our method which use depth estimation method for dense RGB SLAM in large scenes. However, the memory and storage usage are intensive in these GS-based SLAM systems, which makes them difficult to use in large-scale and long-sequence scenarios.

\section{Method}
The pipeline of our system is shown in Fig.~\ref{fig:system}. The inputs of this framework are RGB frames and 3D points $\{I_i, P_i\}_{i=1}^M$ with known camera intrinsics $K \in R_{3
\times3}$. Our model predicts camera poses $\{R_i|t_i\}^M_{i=1}$, color $\mathbf{c}$, and a structured 3D Gaussian scene representation. The system consists of three main modules: (i) Voxel-based progressive scene representation (Sec.~\ref{Sec: mapping}), (ii) 2D-3D fusion camera tracking (Sec.~\ref{Sec: tracking}), (iii) 3D Gaussian-based loop closure (Sec.~\ref{Sec:loop}). The network is incrementally updated with the system operation.

\subsection{Voxel-based Progressive Scene Representation}
\label{Sec: mapping}
Recently, existing methods only focus on small room scenarios and have difficulties in large-scale scenes due to their redundant representation and the cumulative growth in the number of ellipsoids. To address this issue, we design a novel sequentialized voxel-based progressive 3D Gaussian with multiple submaps for compact and efficient scene representation. We utilize the scene structure prior to guide the distribution of Gaussians to remove unnecessary 3D Gaussian ellipsoids, maintaining a low compute cost while avoiding unrestricted growth as the scene expands.

\noindent\textbf{Sequentialized Multi-Resolution Voxel-based Scene Representation} Although there are some voxel-based mapping methods, such as ~\cite{scaffoldgs}, the existing method is not well-suited to the incremental nature of SLAM systems. To better accommodate this characteristic, we further reformulated our approach into a sequentialized multi-resolution voxel-based mapping framework. When a keyframe arrives, the regions observed by this frame is voxelized with the point cloud $\boldsymbol{P}_i\in \mathcal{R}^{N\times3}$, $i$ denotes the ID of the current frame. We use $\mathbf{V_i}\in \mathcal{R}^{N'\times3}$ to denote voxel centers. The center of each voxel is initialized as an anchor point $\textbf{x}_i^a\in \mathcal{R}^3$. Each anchor is characterized by its attributes $\mathcal{A}_i=\left\{\boldsymbol{f}_i^a \in \mathbb{R}^{32}, \boldsymbol{l_i} \in \mathcal{R}^3,\mathcal{O}_i\in \mathcal{R}^{k\times3}\right\}$, where each component represents the anchor feature, scaling, and offsets, respectively. Then, we derive 3D Gaussians attributes from anchor points. The attributes of a neural Gaussian are defined as: position $\mu_i \in \mathcal{R}^3$, opacity $\alpha_i \in \mathcal{R}$, quaternion $q_i\in \mathcal{R}^4$, scaling $s_i\in \mathcal{R}^3$, and color $c_i \in \mathcal{R}^3$. The positions of the corresponding $k$ 3D Gaussians of current frame $I_i$ are calculated as:
\begin{equation}
    \{ \mu_i^m\}_{m=0}^{k-1} = \mathbf{x}_i^a+ \{ \mathcal{O}_i^m\}_{m=0}^{k-1} \cdot \boldsymbol{l}_i
\end{equation}
where $\{ \mathcal{O}_i^m\}_{m=0}^{k-1} \in \mathcal{R}^{k\times3}$ are the learnable offsets and $\boldsymbol{l}_i$ is the scaling factor associated with anchor. Then, the attributes of the Gaussians are decoded from the anchor feature $\boldsymbol{f}^a$, the viewing distance $\mathbf{\delta}_i$,  direction$d_i$ through individual MLPs:
\begin{equation}
    \{ \boldsymbol{f}_i^a, \delta_i, d_i \} \mapsto 
    \{\{ \alpha_i^m\}_{m=0}^{k-1}, \{ q_i^m\}_{m=0}^{k-1} , \{ s_i^m\}_{m=0}^{k-1} ,  \{ c_i^m\}_{m=0}^{k-1}  \} 
\end{equation}
\begin{equation}
    \delta_i=\left\|\mathbf{x}^a_i-\mathbf{x}^c_i\right\|_2, \quad d_i=\frac{\mathbf{x}_i^a-\mathbf{x}_i^c}{\left\|\mathbf{x}_i^a-\mathbf{x}_i^c\right\|_2}
\end{equation}
where $\mathbf{x}^c_i$ denotes camera position of current frame $I_i$. The core MLPs include the opacity MLP , the color MLP  and the covariance
MLP. All of these F* are implemented in a LINEAR $\mapsto$ RELU $\mapsto$ LINEAR style with the hidden dimension
of 32. Each branch’s output is activated with a head layer. For opacity, the output is activated by Tanh, where value 0 serves as a natural threshold for selecting valid samples
and the final valid values can cover the full range of [0,1). For color, we activate the output with Sigmoid function:
\begin{equation}
    \left\{c_0, \ldots, c_{k-1}\right\}=\operatorname{Sigmoid}\left(F_c\right)
\end{equation} which canstrains the color into a range of (0,1). For rotation, we follow 3D-GS and activate it with a
normalization to obtain a valid quaternion. For scaling, we adjust the base scaling of each anchor
with the MLP output as follows:
\begin{equation}
    \left\{s_0, \ldots, s_{k-1}\right\}=\operatorname{Sigmoid}\left(F_s\right) \cdot s_v
\end{equation}

\begin{figure*}[h]

  \centering
   \includegraphics[width=\linewidth]{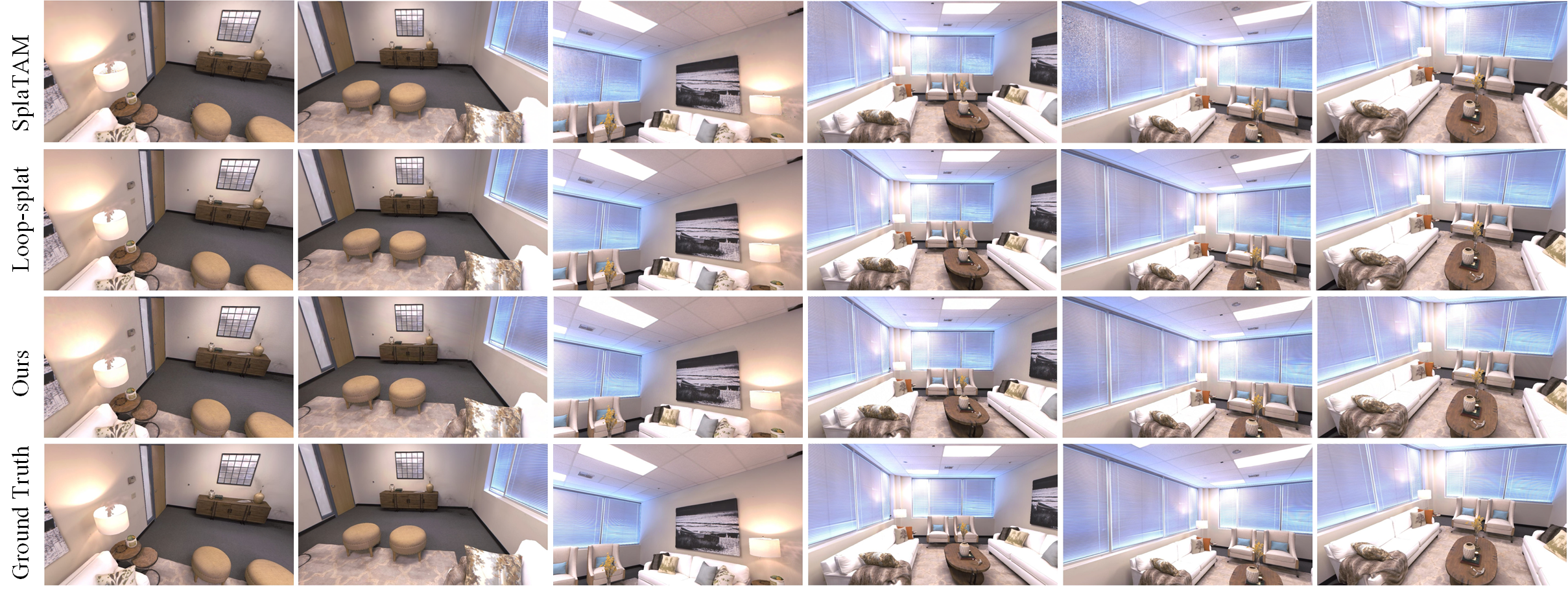}
    \caption{Rendering Performance on Replica dataset~\cite{replica} compared with SplaTAM~\cite{splatam} and Loop-splat~\cite{loopsplat}.}
   \label{fig:replica1}
\end{figure*}

In order to improve the efficiency, we use multiple levels voxel size based on camera distance, from fine to coarse.   All attributes are decoded in a single pass. We achieve online reconstruction through the sequentialized voxel mapping method, where new voxels are continuously initialized, registered, aligned, and iteratively updated, ultimately enabling the reconstruction of the entire sequence.

\noindent\textbf{Progressive Voxel-based Scene Representation}
In order to improve the scalability of scene representation in large-scale scenes, we propose a progressive mapping method that uses multiple submaps to represent the scene to avoid the cumulative growth of 3D Gaussian ellipsoids, similar with. Each submap covers several keyframes that observe it and is represented by a voxelized scene representation.
\begin{equation}
\{I_i,P_i\}_{i=1}^M\mapsto \{\mathrm{M}^1_{\theta_1},\mathrm{M}^2_{\theta_2},\dots ,\mathrm{M}^n_{\theta_n}\}\mapsto\{\mathbf{c},\mathbf{\alpha}\}
\end{equation}
where $\mathrm{M}^n_{\theta_n}$ denotes the $n$-th local submap. Starting with the first keyframe, each submap models a specific region. Whenever the estimated camera pose trajectory leaves the space of the submap, we dynamically allocate a new local scene representation trained with a small set of frames. Subsequently, we progressively introduce additional local frames to the optimization. We use the distance threshold $d$ and the rotation threshold $\omega$ to trigger the initialization of a new submap.

\noindent\textbf{Voxel-Based Submap Expansion and Activation}
Keyframes are selected at fixed intervals for the submap, and we define the first keyframe of the submap as the anchor frame. Every new keyframe adds new anchor points and 3D Gaussians to the active submap for the newly observed parts of the scene. At the beginning of each submap, we first compute a posed point cloud from the input, and then sample $M_k$ points from the regions where the accumulated $\alpha$ is below a threshold or where significant depth discrepancies occur. These points are voxelized and initialized as anchor points. New 3D Gaussian anchors are added to the current submap only if there is no existing 3D Gaussian mean within a radius $\rho$.
Then, for each voxel, we compute the averaged gradients of the included neural Gaussians when we move to the bound of the submap, denoted $\nabla_g$. If the $\nabla_g > \tau_g$, we grow new anchor points in these voxels. To remove trivial anchors, we accumulate the opacity values of their associated neural Gaussians in the submap. To avoid redundant submap creation, we assign the current frame to the most relevant submap by measuring its distance to the anchor frames of all existing submaps.

\noindent\textbf{Online Submap Optimization } In the mapping thread, we optimize the scene representation with the rendering loss. The rendering loss consists of four components:
\begin{equation}
\mathcal{L}_m=\mathcal{L}_c+\mathcal{L}_{\mathrm{SSIM}}+\lambda_d\mathcal{L}_d+\lambda_{\mathrm{vol}} \mathcal{L}_{\mathrm{vol}}
\end{equation}

where $\mathcal{L}_c$ and $\mathcal{L}_{\mathrm{SSIM}}$ are the color losses, and $\mathcal{L}_d$ is the depth loss computed using L1 distance. The SSIM color loss and regularization loss are defined as:
\begin{equation}
    \mathcal{L}_{\mathrm{SSIM}}=\left(1-\lambda_{S S I M}\right) \cdot|\hat{I}-I|_1+\lambda_{S S I M}(1-\operatorname{SSIM}(\hat{I}, I))
\end{equation}
\begin{equation}
\mathcal{L}_{\mathrm{vol}}=\sum_{i=1}^{N_{\mathrm{ng}}} \operatorname{Prod}\left(s_i\right)
\end{equation}
where $N_{\mathrm{ng}}$ denotes the number of neural Gaussians in the submap. $\operatorname{Prod}(\cdot)$ denotes the product of the values of a vector. The volume regularization term promotes small neural Gaussians with minimal overlap. We optimize the parameters of the two relevant submaps only in the overlapping regions. In other areas, only a single submap’s parameters are maintained, while non-essential submaps can be deactivated.  Notably, the number of submap parameters and Gaussian ellipsoids is O(N). Thus, our method significantly reduces online memory consumption, which is crucial for SLAM systems, as existing approaches often encounter GPU out-of-memory issues in large-scale environments. 
\subsection{2D-3D Fusion Camera Tracking}
\label{Sec: tracking}

Most existing 3DGS methods rely on single-modality supervision, typically using rendering loss. However, our findings indicate that 2D photometric loss performs better in indoor and simulated environments, whereas 3D geometric information are more beneficial for outdoor scene reconstruction with motion blur and exposure change.
To this end, we design a coarse-to-fine pose estimation, utilizing both 3D Gaussian information and 2D rendering information for both indoor and large-scale outdoor scenes. In coarse-level optimization, we use 2D photometric information $\mathcal{L}_c,\mathcal{L}_d$ to optimize the camera pose, which provides a good initial estimate. In the fine-level stage, we leverage 3D geometric information to further refine the pose.
However, directly applying 2D-3D optimization is not always suitable for diverse indoor and outdoor environments. To address this, we design a dynamic adaptive method to dynamically adjust the reliance on 2D/3D information during tracking. We assess the quality of rendering results with rendering loss $\mathcal{L}_c,\mathcal{L}_d$ and input frames $I_i$. If the 2D information quality lower than the threshold $\zeta$, we will discard the coarse-level optimization and directly employ 3D Gaussian-based voxel ICP for pose refinement. This strategy significantly improves robustness in large-scale outdoor scenes, where visual inputs may suffer from illumination changes, exposure, and motion blur, enabling the system to rely more heavily on 3D geometric information when necessary.

In the fine-level optimization, we incorporate a voxel-based 3D Gaussian ICP method, inspired by\cite{kissicp}. 
We perform frame-to-map registration with the voxel-based 3D Gaussian map and adopt a double downsampling strategy, which retains only a single original point per voxel. 
Compared to most voxelization strategies that select the center of each voxel for downsampling, we find it more advantageous to retain the original point coordinates during the mapping process, which can avoid discretization errors. For the input points $\mathcal{S}$, we transform
it into the global coordinate frame using the previous pose
estimate $\{R_{i-1},t_{i-1}\}$ and predicted relative pose$\{ R_{\text{pred}},t_{\text{pred}} \}$, resulting in the source points:
\begin{equation}
\mathcal{S}=\left\{\boldsymbol{s_i}=\{R_{i-1},t_{i-1}\} \{ R_{\text{pred}},t_{\text{pred}} \}\boldsymbol{p} \mid \boldsymbol{p} \in \boldsymbol{P}\right\} .
\end{equation}
where $\boldsymbol{P}$ denotes the scan in the local frame.
We obtain a set of correspondences between the point cloud $\mathcal{S}$ and the local map $\mathcal{M}=\left\{\boldsymbol{m}_i \mid \boldsymbol{m}_i \in \mathbb{R}^3\right\}$ through nearest neighbor(NN) search over
the 3D Gaussian voxel map considering only correspondences
with a point-to-point distance below threshold. To compute the current
pose registration, we perform a robust optimization minimizing the residuals:
\begin{equation}
\Delta \{R,t\}_{\text {est }}=\underset{\{R,t\}}{\operatorname{argmin}} \sum_{(s, q) \in \mathcal{C}\left(\tau_t\right)} \rho\left(\|\{R,t\} \boldsymbol{s}-\boldsymbol{m}\|_2\right) 
\end{equation}
\begin{equation}
    \rho(e)=\frac{e^2 / 2}{\sigma_t / 3+e^2}
\end{equation}
where $\mathcal{C}\left(\tau_t\right)$ is the set of nearest neighbor correspondences with a distance smaller than $\tau_t$ and $\rho$ is the Geman-McClure robust kernel. The scale factor $\sigma_t$ of the kernel is adapted online.
Furthermore, we use a hash table to represent the voxel structure, which offers memory-efficient storage and enables fast nearest neighbor search.

\begin{table*}[htbp]
  \centering
  \begin{minipage}[c]{0.48\linewidth}
    \centering
    \scalebox{0.95}{
\setlength{\tabcolsep}{0.7mm}{
    \begin{tabular}{lcccccccccc}
\toprule
Methods       & LC                      & Rm0                     & Rm1                     & Rm2                     & Off0 & Off1 & Off2 & Off3 & Off4 & Avg.                     \\ \midrule
\multicolumn{11}{l}{\cellcolor[HTML]{EEEEEE}{\textit{NeRF-based Methods}}} \\
NICE-SLAM\cite{niceslam}     & \noo                     & 0.93                     & 1.28                     & 1.07                     & 0.93  & 1.04  & 1.08  & 1.12  & 1.13  & 1.07                     \\
ESLAM\cite{eslam}         & \noo                     & 0.72                     & 0.68                     & 0.54                     & 0.55  & 0.55  & 0.60  & 0.70  & 0.65  & 0.65                     \\
Co-SLAM\cite{coslam}       & \noo                     & 0.96                     & 1.02                     & 0.95                     & 0.83  & 0.87  & 0.94  & 1.01  & 0.93  & 0.95                     \\
Point-SLAM\cite{pointslam}    & \yes                     & 0.52                     & 0.38                     & 0.30                     & 0.50  & 0.44  & 1.26  & 0.77  & 0.57  & 0.59                     \\
PLGSLAM\cite{plgslam}       & \noo                     & 0.64                     & 0.65                     & 0.49                     & 0.51  & 0.52  & 0.54  & 0.65  & 0.57  & 0.57 \\
Loopy-SLAM\cite{loopyslam}    & \yes  & \cellcolor{tabsecond}0.27 & 0.25  & 0.29  & \cellcolor{tabthird}0.29  & 0.41 & 0.31  & 0.24  & 0.34  & \cellcolor{tabthird}0.31  \\ 
 \hdashline
\multicolumn{11}{l}{\cellcolor[HTML]{EEEEEE}{\textit{3DGS-based Methods}}} \\
SplaTAM\cite{splatam}       & \noo                     & 0.33                     & 0.44                     & 0.33                     & 0.51  & \cellcolor{tabthird}0.29  & \cellcolor{tabthird}0.29  & 0.35  & 0.75  & 0.41                     \\
MonoGS\cite{monogs}        & \noo                     & 0.35                     & \cellcolor{tabthird}0.24                     & 0.32                     & 0.38  & 0.22  & \cellcolor{tabsecond}0.26  & \cellcolor{tabfirst}0.16  & 0.82  & 0.34                     \\
Gaussian-SLAM\cite{gaussianslam} & \noo                     & 0.29                     & 0.29                     & \cellcolor{tabthird}0.23                     & 0.38  & 0.24  & 0.42  & 0.31  & \cellcolor{tabthird}0.36  & 0.32                     \\ Photo-SLAM\cite{photoslam}       & \noo                     & 0.65                     & 0.65                     & 0.50                     & 0.53  & 0.52  & 0.55  & 0.66  & 0.58  & 0.58 \\
Loop-Splat\cite{loopsplat}    & \yes                     & \cellcolor{tabthird}0.28                     & \cellcolor{tabsecond}0.22                     & \cellcolor{tabsecond}0.17                     & \cellcolor{tabsecond}0.22  & \cellcolor{tabfirst}0.15  & 0.49  & \cellcolor{tabthird}0.20  & \cellcolor{tabthird}0.30  & \cellcolor{tabthird}0.26                     \\
Ours          & \yes & \cellcolor{tabfirst}0.26 & \cellcolor{tabfirst}0.21 & \cellcolor{tabfirst}0.16 & \cellcolor{tabfirst}0.21  & \cellcolor{tabfirst}0.15  & \cellcolor{tabfirst}0.25  & \cellcolor{tabsecond}0.18  & \cellcolor{tabfirst}0.28  & \cellcolor{tabfirst}0.21 \\ \bottomrule
\end{tabular}}}
\caption{\textbf{Camera Tracking Performance on Replica dataset~\cite{replica}}. We use ATE RMSE (cm) as the metric. Best results are highlighted as \colorbox{tabfirst}{first}, \colorbox{tabsecond}{second}, and \colorbox{tabthird}{third}.}

\label{tab:track1}
  \end{minipage}%
  \hfill
  \begin{minipage}[c]{0.48\linewidth}
    \centering
    \scalebox{0.95}{
\setlength{\tabcolsep}{0.8mm}{
\begin{tabular}{lccccccccc}
\toprule
Methods       & 0000   & 0059   & 0106  & 0169  & 0181  & 0207  & 0054   & 0233  & Avg. \\ \midrule
\multicolumn{10}{l}{\cellcolor[HTML]{EEEEEE}{\textit{NeRF-based Methods}}} \\
NICE-SLAM\cite{niceslam}     & 12.3 & 14.2 & 7.9  & 10.9 & 13.6 & 6.8  & 20.9 & 9.4  & 13.2 \\
ESLAM\cite{eslam}         & 7.7  & 8.6  & 7.8  & \cellcolor{tabthird}6.5  & 9.4  & \cellcolor{tabsecond}6.2  & 36.5 & \cellcolor{tabsecond}4.6  & 10.7 \\
Co-SLAM\cite{coslam}       & 7.2  & 11.4 & 9.5  & \cellcolor{tabfirst}5.9  & 11.9 & 7.3  & -    & -    & -    \\
Point-SLAM\cite{pointslam}    & 10.4 & 7.9  & 8.8  & 22.2 & 14.9 & 9.7  & 28.2 & 6.3  & 14.4 \\
GO-SLAM\cite{goslam}       & \cellcolor{tabsecond}5.7 & \cellcolor{tabthird}7.5 & \cellcolor{tabfirst}7.0 & 7.8 & \cellcolor{tabfirst}6.8 & 6.7 & \cellcolor{tabfirst}8.8 & 4.9 & \cellcolor{tabfirst}6.8 \\
PLGSLAM\cite{plgslam}      & 7.3  & 8.2  & \cellcolor{tabthird}7.4  & \cellcolor{tabsecond}6.3  & 9.2  & \cellcolor{tabfirst}5.8  & 30.8 & \cellcolor{tabfirst}4.2  & 9.9  \\
Loopy-SLAM\cite{loopyslam}    & \cellcolor{tabfirst}4.9  & 7.7  & 8.5  & 7.7  & 10.6 & 7.9  & \cellcolor{tabthird}14.5  & 5.2  & \cellcolor{tabthird}7.7  \\
\multicolumn{10}{l}{\cellcolor[HTML]{EEEEEE}{\textit{3DGS-based Methods}}} \\
SplaTAM\cite{splatam}       & 12.8 & 10.1 & 17.7 & 12.1 & 11.1 & 7.5  & 56.8 & 4.8  & 16.6 \\
MonoGS\cite{monogs}        & 9.8  & 32.1 & 8.9  & 10.7 & 21.8 & 7.9  & 17.5 & 12.4 & 15.2 \\
Gaussian-SLAM\cite{gaussianslam} & 21.2 & 12.8 & 13.5 & 16.3 & 21.0 & 14.3 & 37.1 & 11.1 & 18.4 \\
Loop-Splat\cite{loopsplat}    & 6.2 & \cellcolor{tabfirst}7.2 & \cellcolor{tabthird}7.4 & 10.8 & \cellcolor{tabthird}8.5 & 6.7 & 16.3 & 4.8 & 8.4  \\
Ours          & \cellcolor{tabsecond}5.7 & \cellcolor{tabfirst}7.2 & \cellcolor{tabsecond}7.1 & 9.1 & \cellcolor{tabsecond}7.5 & \cellcolor{tabsecond}6.2 & \cellcolor{tabsecond}13.9 & \cellcolor{tabsecond}4.6 & \cellcolor{tabsecond}7.6 \\ \bottomrule
\end{tabular}}}
\caption{\textbf{Camera Tracking Performance on ScanNet dataset~\cite{scannet}}. We use ATE RMSE (cm) as the metric. We evaluate on eight sequences following the experimental settings of previous methods.}
\label{tab:track2}
  \end{minipage}

\end{table*}

\begin{figure*}[t] 
\center
\includegraphics[width=\linewidth]{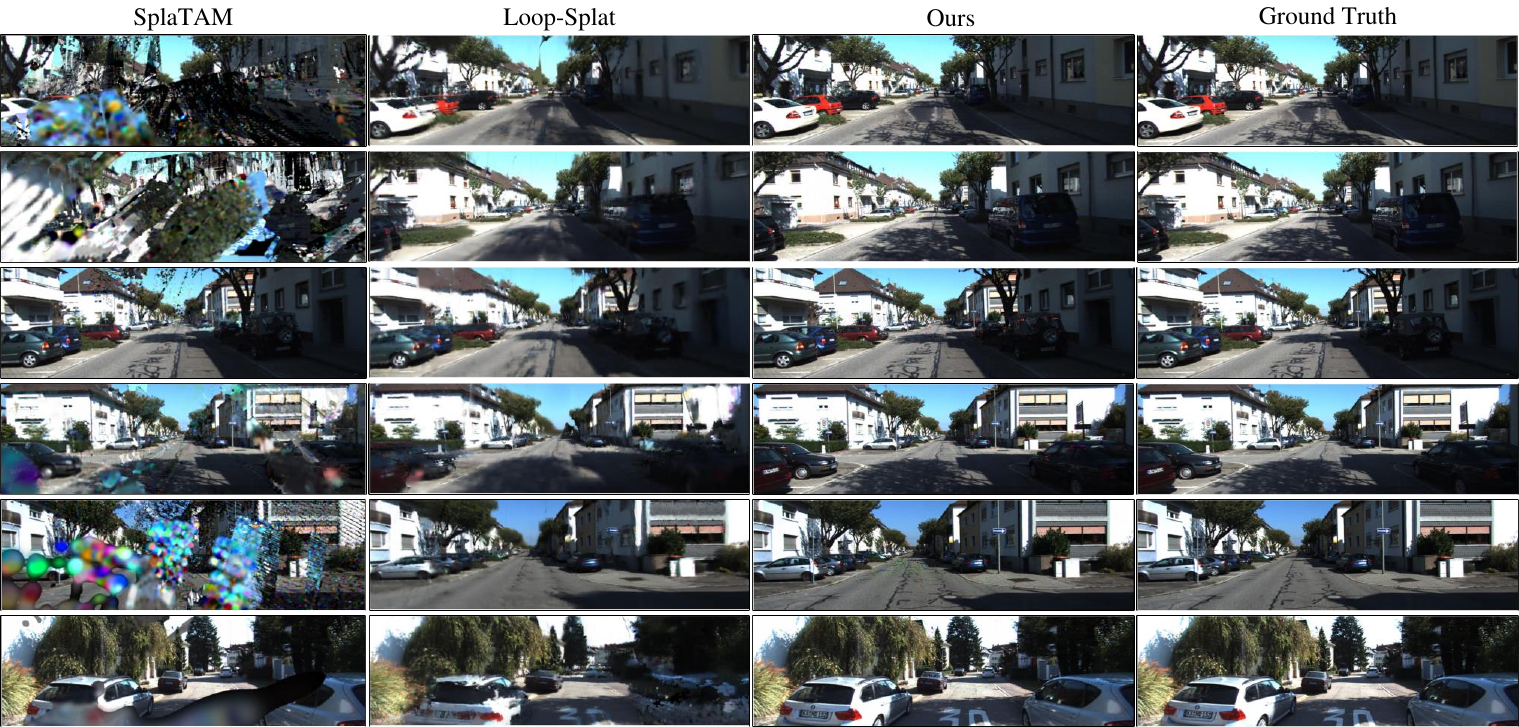}

\caption{Qualitative comparison between our proposed method and existing SOTA methods: SplaTAM~\cite{splatam} and Loop-Splat~\cite{loopsplat}. We demonstrate RGB image rendering results on the KITTI odometry dataset~\cite{kitti}. Our method shows improved rendering quality compared to these existing methods.}

\label{fig:kitti1}
\end{figure*}

\begin{figure*}[t] 
\center
\includegraphics[width=1.0\textwidth]{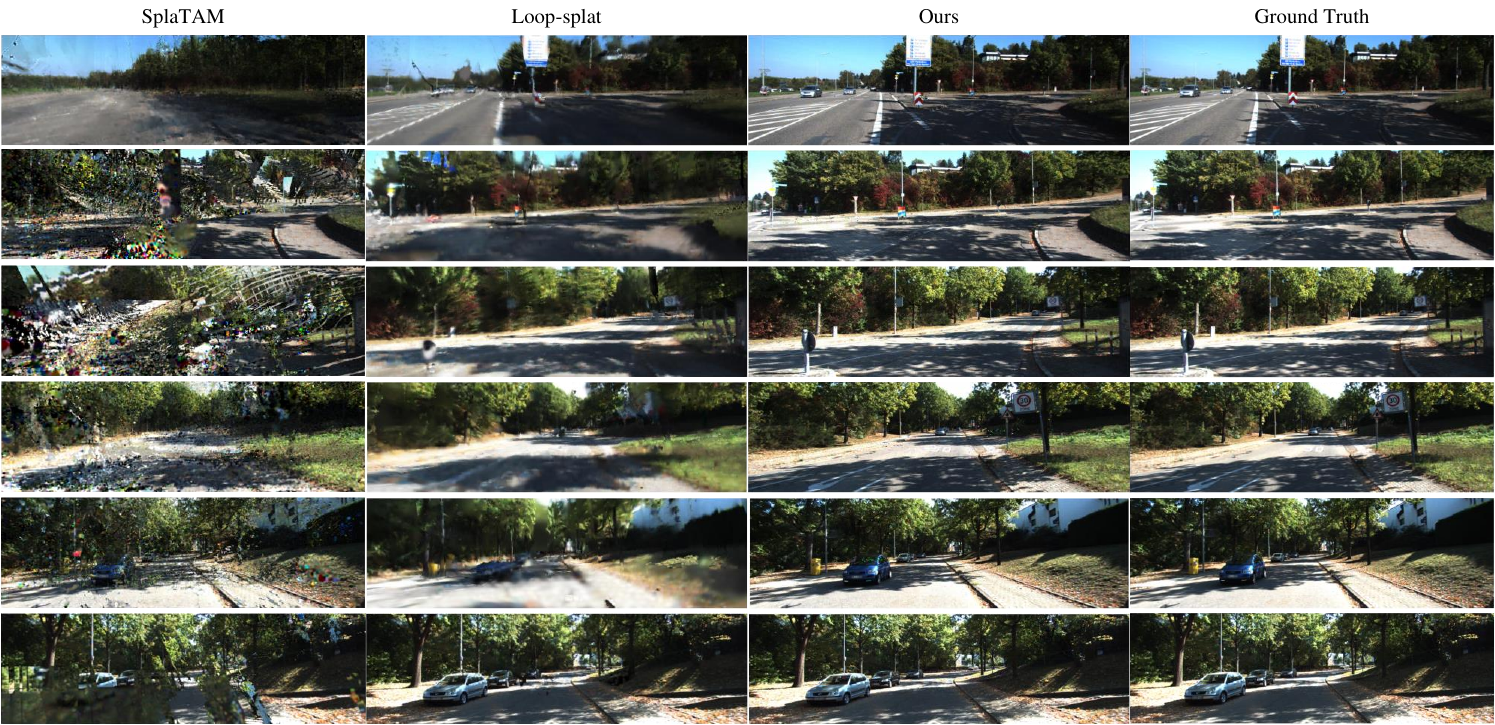}
\caption{Visualization of RGB image rendering results on the KITTI odometry dataset sequence 02. Qualitative comparison between our proposed method and existing SOTA methods: SplaTAM~\cite{splatam} and Loop-Splat~\cite{loopsplat}. We demonstrate RGB image rendering results on the KITTI odometry dataset. Our method shows improved rendering quality compared
to these existing methods.
}
\label{fig:kitti2}
\end{figure*}

\begin{figure*}[t] 
\center
\includegraphics[width=1.0\textwidth]{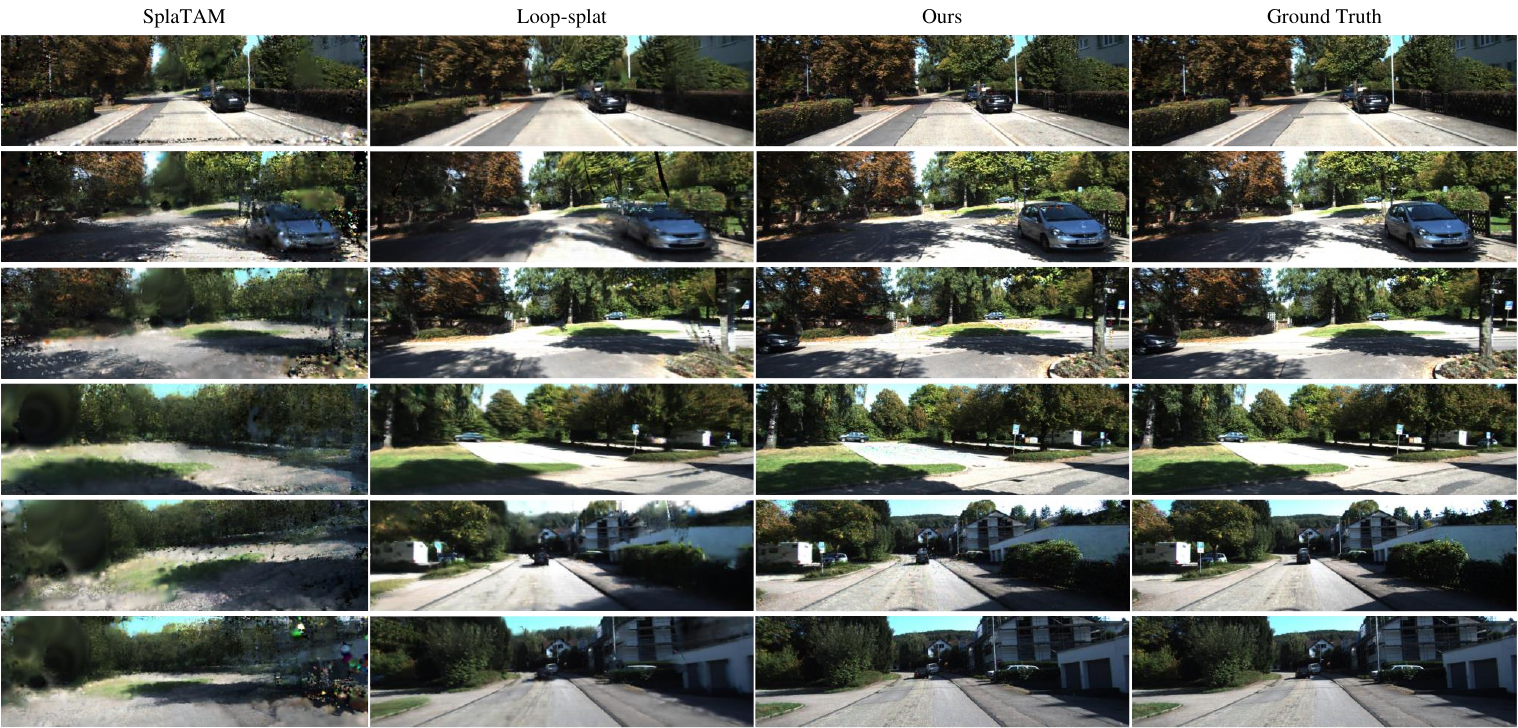}
\caption{Visualization of RGB image rendering results on the KITTI odometry dataset sequence 03. Qualitative comparison between our proposed method and existing SOTA methods: SplaTAM~\cite{splatam} and Loop-Splat~\cite{loopsplat}. We demonstrate RGB image rendering results on the KITTI odometry dataset. Our method shows improved rendering quality compared
to these existing methods.
}
\label{fig:kitti3}
\end{figure*}

\begin{figure*}[h]

  \centering
   \includegraphics[width=\linewidth]{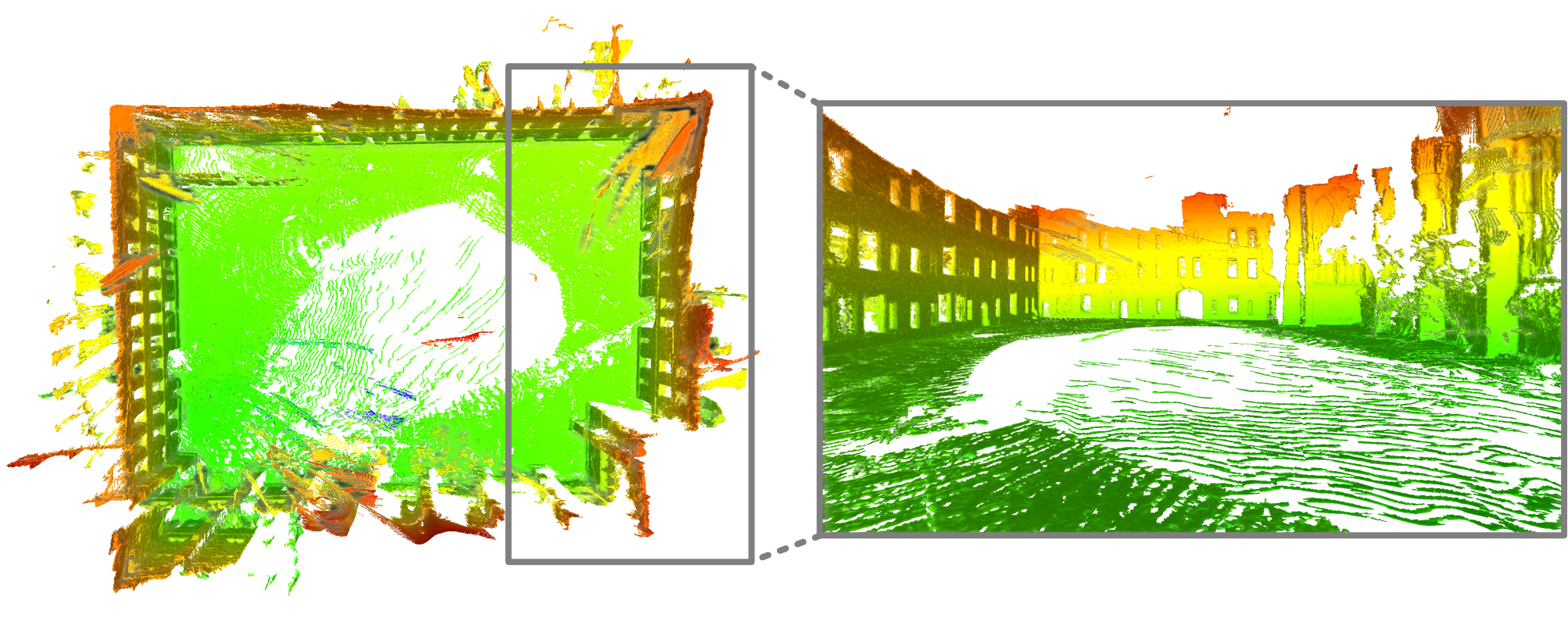}
    \caption{Mesh reconstruction results on Newer College dataset~\cite{newer}.}
   \label{fig:newer}
\end{figure*}
\subsection{2D-3D Gaussian Loop Closure }
\label{Sec:loop}
In large-scale environments, pose drift accumulates significantly, posing a major challenge for existing methods. Furthermore, the fusion of multiple submaps also requires alignment to achieve global consistency. We incorporate a novel loop closure method into our framework based on 3D Gaussians to identify pose corrections for past submaps and keyframes. Although some existing GS-SLAM methods~\cite{loopsplat} have incorporated loop closure, we improve the loop closure optimization by combining 2D photometric loss with voxel-based ICP, which allows better adaptation to domain differences between indoor and outdoor environments. Furthermore, we design an online distillation methods for submap fusion upon loop closure detection, enabling the integration of information from different submaps to achieve global consistency.

\noindent\textbf{Loop Detection and Correction}
In order to detect when the robot revisits the same place, we extract the visual descriptor with a lightweight method Netvlad~\cite{netvlad}. We then compute the cosine similarities of the descriptor across different submap $\alpha$, $\beta$. 
Upon detecting a loop, we construct pose optimization graph with the loop closure constraints. We utilize both 2D rendering loss and voxel based ICP as loop closure constraints to achieve better adaptation to domain differences. We obtain a set of pose corrections $\{R_i|t_i\}_M$ from PGO, where $M$ denotes the correction for submap $M$. Then we update both the camera pose and attributes of 3D Gaussians:
\begin{equation}
    \{R_i|t_i\}_M \cdot \{R_i|t_i\} \mapsto \{R_i|t_i\}, \quad \{R_i|t_i\}_M \{ q_i\}_{i=0}^{k-1}  \mapsto  \{ q_i\}_{i=0}^{k-1} 
\end{equation}
where $\{ q_i\}_{i=0}^{k-1}$ denotes the quaternion of 3D Gaussians anchor in submap $M$. 

\noindent\textbf{Submap Fusion}
After aligning the relative poses between submaps, we further propose a submap fusion method to integrate information across different submaps and ensure global consistency.
Specifically, for two submaps where a loop closure is detected, we propose an online distillation approach.
we identify highly similar image pairs between the two submaps as overlapping regions, and then construct a distillation loss to merge the corresponding 3D Gaussians from the two submaps effectively and improve spatial correlation. We use the poses of the overlapping region keyframe $\{KF\}_{i=1}^m$ to render the RGB and depth images.
 \begin{equation}
     \mathcal{L}_{\alpha-\beta}=\frac{1}{m} \sum_{\alpha,\beta \in \{KF \}}^m\left( \left(\hat{\mathbf{c}}_\alpha-\mathbf{\hat{c}}_\beta \right)^2 + \left(\mathbf{\hat{d_\alpha}}-\mathbf{\hat{d}_\beta}\right)^2 
       \right)
     \end{equation}
    where $\hat{\mathbf{c}}_\alpha$, $\hat{\mathbf{c}}_\beta$, and $\hat{\mathbf{d}}_\alpha$ and $\hat{\mathbf{d}}_\beta$ denotes the rendered color and depth images from different submap $\alpha$ and $\beta$ from different networks.

\begin{table}[htbp]
  \centering
    \scalebox{0.95}{
\setlength{\tabcolsep}{1.5mm}{
\begin{tabular}{lccc}
\hline
Methods       & Accuracy & Completion & Comp. Ratio  \\ \hline
SplaTAM~\cite{splatam}       & 2.74     & 4.02       & 84.61     \\
MonoGS~\cite{monogs}        & 3.16     & 4.45       & 81.52     \\
Gaussian-SLAM~\cite{gaussianslam} & \cellcolor{tabsecond}2.53     & \cellcolor{tabthird}3.77       & \cellcolor{tabthird}84.65     \\
Splat-SLAM~\cite{splatslam}    & \cellcolor{tabthird}2.49     & \cellcolor{tabsecond}3.68       & \cellcolor{tabsecond}84.79     \\
Ours          & \cellcolor{tabfirst}2.47     & \cellcolor{tabfirst}3.64       & \cellcolor{tabfirst}85.75     \\ \hline
\end{tabular}}}

\caption{\textbf{Scene Reconstruction Performance on Replica dataset~\cite{replica}.} We use accuracy [cm], completion [cm] and completion ratio (\%) as the metrics for mesh evaluation.}

\label{tab:mesh}
\end{table}

\begin{figure*}[ht]
    \centering
    
    \includegraphics[width=\linewidth]{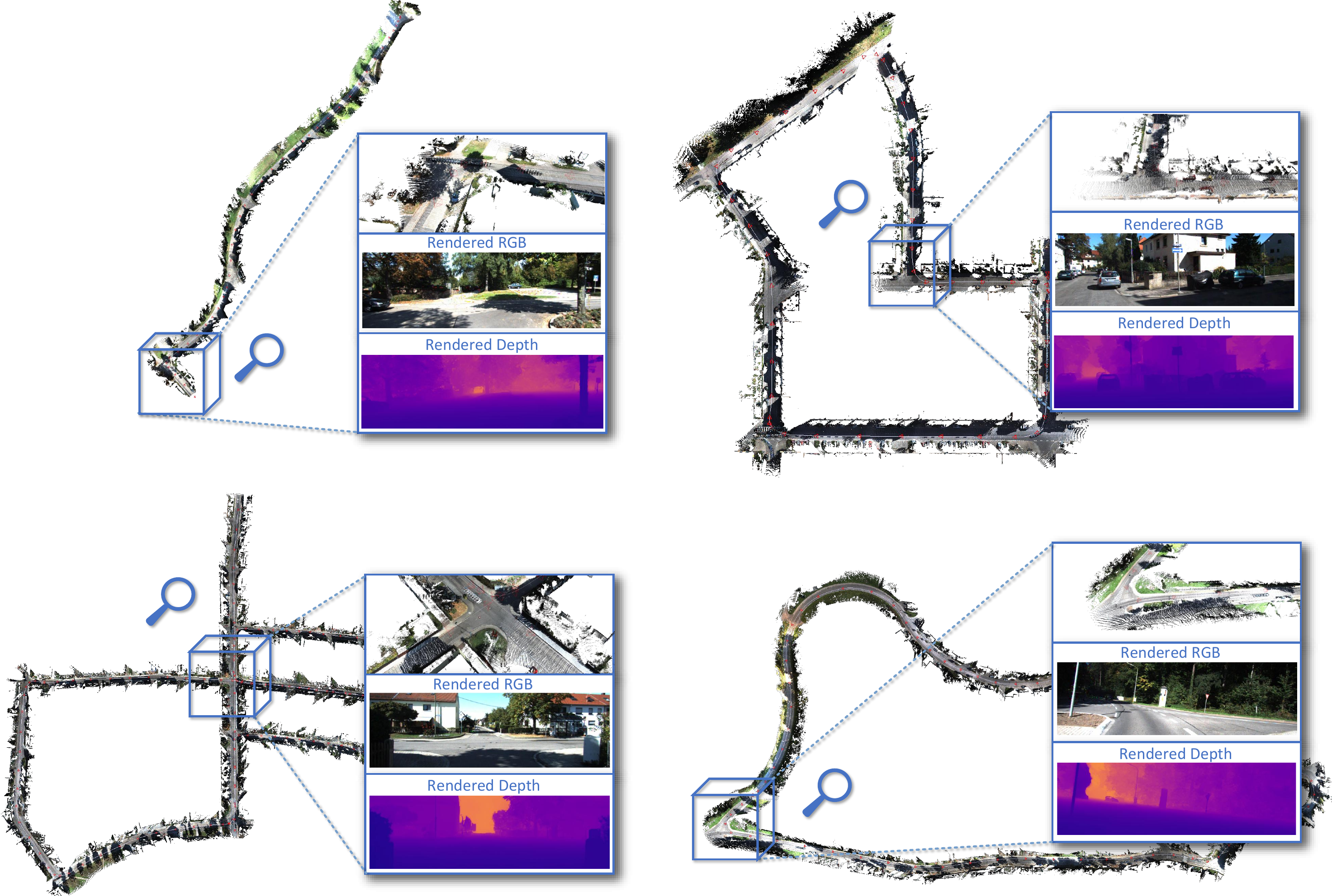}

    \caption{\textbf{Extensive Experiments on KITTI datasets~\cite{kitti}.} Our system is a large-scale SLAM framework, with voxel-based progressive 3D Gaussian representation, 2D-3D fusion camera tracking, and 3D Gaussian loop closure. Our framework takes color images and 3D point clouds as input. Our method can achieve accurate and efficient scene reconstruction, camera tracking, and global map generation.}

    \label{fig:track1}
\end{figure*}

    



\begin{table}[ht]
        \centering
\scalebox{0.95}{
\setlength{\tabcolsep}{1.2mm}{
\begin{tabular}{lcccc}
\hline
\multirow{2}{*}{Methods}           & \multicolumn{2}{c}{VKITTI2~\cite{vkitti2}}       & \multicolumn{2}{c}{KITTI~\cite{kitti}}         \\ \cline{2-5} 
                                   & PSNR $\uparrow$ & SSIM $\uparrow$ & PSNR $\uparrow$ & SSIM $\uparrow$ \\ \hline
\multicolumn{1}{l|}{GO-SLAM~\cite{goslam}}       & \cellcolor{tabthird}{19.29}           & \cellcolor{tabthird}{0.73}            & \cellcolor{tabthird}{15.71}           & \cellcolor{tabthird}{0.51}            \\
\multicolumn{1}{l|}{SplaTAM~\cite{splatam}}       & 18.29           & 0.69            & 14.68           & 0.48            \\
\multicolumn{1}{l|}{MonoGS~\cite{monogs}}        & 18.45           & 0.70            & 14.73           & 0.49            \\
\multicolumn{1}{l|}{Gaussian-SLAM~\cite{gaussianslam}} & 18.78           & 0.69            & 14.78           & 0.50            \\
\multicolumn{1}{l|}{Loop-Splat~\cite{loopsplat}}    & \cellcolor{tabsecond}{19.47}           & \cellcolor{tabsecond}{0.76}            & \cellcolor{tabsecond}{16.77}           & \cellcolor{tabsecond}{0.71}           \\
\multicolumn{1}{l|}{Ours}          & \cellcolor{tabfirst}{25.45}           & \cellcolor{tabfirst}{0.84}            & \cellcolor{tabfirst}{21.37}           & \cellcolor{tabfirst}{0.81}            \\ \hline
\end{tabular}}}
  \caption{\textbf{Scene Reconstruction Performance comparison on KITTI~\cite{kitti} and VKITTI 2~\cite{vkitti2} datasets. } We use PSNR and SSIM as the metrics.  }
  \label{tab:kittirender}
\end{table}

\begin{table*}[]
    \centering
   \centering
    \scalebox{0.95}{
\setlength{\tabcolsep}{2mm}{
\begin{tabular}{llccccccccc}
\toprule
Method                                                                 & Metrics & Room0   & Room1   & Room2   & Office0  & Office1  & Office2  & Office3  & Office4  & Avgerage  \\ \midrule
\multicolumn{11}{l}{\cellcolor[HTML]{EEEEEE}{\textit{NeRF-based Methods}}} \\
\multirow{3}{*}{ESLAM\cite{eslam}}                                                 & Depth  & 0.97  & 1.07  & 1.28  & 0.86  & 1.26  & 1.71  & 1.43  & 1.06  & 1.18  \\
                                                                       & PSNR   & 25.25 & 27.39 & 28.09 & 30.33 & 27.04 & 27.99 & 29.27 & 29.15 & 28.06 \\
                                                                       & SSIM   & 0.874 & 0.891 & 0.935 & 0.934 & 0.910 & 0.942 & 0.953 & 0.948 & 0.923 \\
\multirow{3}{*}{Co-SLAM\cite{coslam}}                                               & Depth  & 1.05  & 0.85  & 2.37  & 1.24  & 1.48  & 1.86  & 1.66  & 1.54  & 1.51  \\
                                                                       & PSNR   & 27.27 & 27.45 & 28.12 & 30.31 & 28.47 & 28.91 & 29.75 & 29.91 & 28.77 \\
                                                                       & SSIM   & 0.898 & 0.905 & 0.937 & 0.946 & 0.923 & 0.952 & 0.961 & 0.957 & 0.934 \\
\multirow{3}{*}{GO-SLAM\cite{goslam}}                                               & Depth  & 2.56  & 1.57  & 2.43  & 1.47  & 1.63  & 2.31  & 1.71  & 1.47  & 4.68  \\
                                                                       & PSNR   & 24.32 & 25.37 & 26.52 & 30.17 & 30.34 & 24.16 & 28.23 & 27.64 & 24.42 \\
                                                                       & SSIM   & 0.854 & 0.871 & 0.905 & 0.904 & 0.893 & 0.912 & 0.927 & 0.918 & 0.898 \\
\multirow{3}{*}{\begin{tabular}[c]{@{}l@{}}Point-\\ SLAM\cite{pointslam}\end{tabular}} & Depth  & 0.53  & \cellcolor{tabsecond}0.22  & \cellcolor{tabsecond}0.46  & \cellcolor{tabsecond}0.30  & 0.57  & \cellcolor{tabfirst}0.49  & \cellcolor{tabsecond}0.51  & 0.46  & \cellcolor{tabsecond}0.44  \\
                                                                       & PSNR   & 32.40 & \cellcolor{tabthird}34.08 & 35.50 & 38.26 & 39.16 & 33.99 & \cellcolor{tabthird}33.48 & 33.49 & 35.17 \\
                                                                       & SSIM   & \cellcolor{tabthird}0.974 & 0.977 & 0.982 & 0.983 & 0.986 & 0.960 & 0.960 & 0.979 & 0.975 \\
\multirow{3}{*}{\begin{tabular}[c]{@{}l@{}}Loopy-\\ SLAM\cite{loopyslam}\end{tabular}} & Depth  & \cellcolor{tabfirst}0.33  & \cellcolor{tabfirst}0.21  & \cellcolor{tabfirst}0.43  & \cellcolor{tabfirst}0.24  & \cellcolor{tabfirst}0.47  & \cellcolor{tabthird}0.62  & \cellcolor{tabfirst}0.37  & \cellcolor{tabfirst}0.26  & \cellcolor{tabfirst}0.37  \\
                                                                       & PSNR   & 32.71 & 34.28 & \cellcolor{tabthird}35.70 & \cellcolor{tabthird}38.39 & 38.91 & \cellcolor{tabthird}34.09 & \cellcolor{tabthird}33.48 & \cellcolor{tabthird}33.79 & \cellcolor{tabthird}35.19 \\
                                                                       & SSIM   & \cellcolor{tabfirst}0.984 & \cellcolor{tabsecond}0.980 & \cellcolor{tabfirst}0.989 & \cellcolor{tabthird}0.985 & \cellcolor{tabfirst}0.991 & \cellcolor{tabthird}0.974 & \cellcolor{tabthird}0.967 & \cellcolor{tabfirst}0.985 & \cellcolor{tabthird}0.980 \\ \midrule
                                                                       \multicolumn{11}{l}{\cellcolor[HTML]{EEEEEE}{\textit{3DGS-based Methods}}} \\
\multirow{3}{*}{SplaTAM\cite{splatam}}                                               & Depth  & 0.43  & 0.38  & 0.54  & 0.44  & 0.66  & 1.05  & 1.60  & 0.68  & 0.72  \\
                                                                       & PSNR   & \cellcolor{tabthird}32.86 & 33.89 & 35.25 & 38.26 & \cellcolor{tabthird}39.17 & 31.97 & 29.70 & 31.81 & 34.11 \\
                                                                       & SSIM   & \cellcolor{tabsecond}0.982 & 0.974 & \cellcolor{tabthird}0.983 & 0.981 & 0.981 & 0.971 & 0.951 & 0.953 & 0.972 \\
\multirow{3}{*}{\begin{tabular}[c]{@{}l@{}}Loop-\\ Splat\cite{loopsplat}\end{tabular}} & Depth  & \cellcolor{tabthird}0.39  & 0.23  & 0.52  & 0.32  & \cellcolor{tabthird}0.51  & 0.63  & 1.09  & \cellcolor{tabthird}0.40  & \cellcolor{tabthird}0.51  \\
                                                                       & PSNR   & \cellcolor{tabsecond}33.07 & \cellcolor{tabfirst}35.32 & \cellcolor{tabsecond}36.16 & \cellcolor{tabfirst}39.12 & \cellcolor{tabfirst}39.81 & \cellcolor{tabsecond}34.67 & \cellcolor{tabsecond}33.93 & \cellcolor{tabsecond}33.98 & \cellcolor{tabsecond}35.75 \\
                                                                       & SSIM   & 0.971 & \cellcolor{tabthird}0.978 & 0.981 & \cellcolor{tabsecond}0.989 & \cellcolor{tabthird}0.988 & \cellcolor{tabfirst}0.981 & \cellcolor{tabsecond}0.987 & \cellcolor{tabthird}0.984 & \cellcolor{tabsecond}0.982 \\
\multirow{3}{*}{Ours}                                                  & Depth  & \cellcolor{tabsecond}0.34  & \cellcolor{tabsecond}0.22  & \cellcolor{tabthird}0.50  & \cellcolor{tabsecond}0.30  & \cellcolor{tabsecond}0.49  & \cellcolor{tabsecond}0.59  & \cellcolor{tabthird}0.98  & \cellcolor{tabsecond}0.38  & \cellcolor{tabthird}0.51  \\
                                                                       & PSNR   & \cellcolor{tabfirst}33.10      & \cellcolor{tabsecond}35.09      & \cellcolor{tabfirst}36.21       & \cellcolor{tabfirst}39.12       & \cellcolor{tabsecond}39.79       & \cellcolor{tabfirst}35.07      & \cellcolor{tabfirst}34.09      & \cellcolor{tabfirst}34.12      & \cellcolor{tabfirst}35.82      \\
                                                                       & SSIM   & 0.973      & \cellcolor{tabfirst}0.981      & \cellcolor{tabsecond}0.985       & \cellcolor{tabfirst}0.991       & \cellcolor{tabfirst}0.991       & \cellcolor{tabfirst}0.981      &\cellcolor{tabfirst}0.988       & \cellcolor{tabfirst}0.985       & \cellcolor{tabfirst}0.984       \\ \bottomrule
\end{tabular}}}

\caption{\textbf{Scene Reconstruction Performance on Replica dataset~\cite{replica}}. We use depth L1, PSNR, SSIM as our metics.}
\label{tab:render}

\end{table*}

\begin{table*}[t]
  \centering
  \scalebox{0.95}{
  \setlength{\tabcolsep}{2.1mm}{
  \begin{tabular}{lclcccccccccccc}
\toprule
Methods                                           & \multicolumn{2}{c}{Map Type}                     & 00   & 01    & 02    & 03   & 04   & 05   & 06   & 07   & 08   & 09   & 10   & Average \\ \midrule
Suma~\cite{suma}                                                                  & \multicolumn{2}{c}{Surfel}                       & \cellcolor{tabthird}2.94 & 13.85 & 8.43  & 0.94 & 0.43 & \cellcolor{tabthird}1.26 & \cellcolor{tabthird}0.47 & 0.54 & 2.87 & 2.95 & 1.37 & 3.61     \\ 
Litamin2~\cite{litamin2}                                                              & \multicolumn{2}{c}{NDT}                          & 5.84 & 15.93 & 10.74 & 0.85 & 0.77 & 2.46 & 0.91 & 0.65 & \cellcolor{tabsecond}2.57 & 2.13 & 1.05 & 4.39     \\ \hline
FLOAM~\cite{floam}                                                                & \multicolumn{2}{c}{\multirow{3}{*}{Point Cloud}} & 5.03 & \cellcolor{tabthird}3.27  & 8.64  & \cellcolor{tabthird}0.74 & 0.35 & 3.43 & 0.53 & 0.63 & \cellcolor{tabsecond}2.57 & 2.14 & 1.08 & 2.84     \\
KISS-ICP~\cite{kissicp}                                                              & \multicolumn{2}{c}{}                             & 3.72 & 10.35 & 7.86  & 2.13 & 0.51 & 1.37 & 0.65 & 0.63 & 3.64 & 2.35 & 1.49 & 3.47     \\
\begin{tabular}[c]{@{}l@{}}Mesh-Loam~\cite{meshloam} \\ (point-to-plane)\end{tabular} & \multicolumn{2}{c}{}                             & 5.54 & 4.09  & \cellcolor{tabthird}7.05  & \cellcolor{tabsecond}0.57 & 1.53 & 1.74 & 8.94 & 0.56 & 3.07 & 1.84 & \cellcolor{tabfirst}0.95 & 3.59     \\ \hline
Puma~\cite{puma}                                                                  & \multicolumn{2}{c}{\multirow{3}{*}{Mesh}}        & 6.64 & 32.61 & 18.55 & 2.25 & 0.94 & 3.34 & 2.41 & 0.93 & 6.34 & 3.94 & 4.43 & 8.23     \\
SLAMesh~\cite{slamesh}                                                               & \multicolumn{2}{c}{}                             & 5.51 & 10.93 & 13.29 & 0.83 & 0.33 & 3.74 & 0.71 & 0.83 & 5.13 & \cellcolor{tabfirst}1.15 & 1.17 & 4.36     \\
Mesh-Loam~\cite{meshloam}                                                             & \multicolumn{2}{c}{}                             & 5.37 & \cellcolor{tabsecond}3.25  & 7.45  & \cellcolor{tabfirst}0.54 & 0.37 & 1.74 & \cellcolor{tabsecond}0.38 & \cellcolor{tabthird}0.45 & 3.35 & 1.74 & \cellcolor{tabfirst}0.95 & \cellcolor{tabthird}2.77     \\ \midrule
NeRF-Loam~\cite{nerfloam}                                                             & \multicolumn{2}{c}{\multirow{3}{*}{NeRF}}        & 8.64 & 20.58 & 7.45  & 1.79 & 0.80 & 5.01 & 2.47 & 0.79 & 4.76 & 3.47 & 3.02 & 5.88     \\
PIN-LO~\cite{pinslam}                                                                & \multicolumn{2}{c}{}                             & 5.84 & 4.37  & 9.35  & 0.85 & \cellcolor{tabthird}0.19 & 1.87 & 0.53 & 0.55 & 3.27 & 2.35 & 0.97 & 3.01     \\
PIN-SLAM~\cite{pinslam}                                                              & \multicolumn{2}{c}{}                             & \cellcolor{tabsecond}1.18 & 3.46  & \cellcolor{tabsecond}2.69  & 0.80 & \cellcolor{tabfirst}0.17 & \cellcolor{tabsecond}0.37 & 0.48 & \cellcolor{tabfirst}0.30 & 2.58 & \cellcolor{tabthird}1.34 & 0.96 & \cellcolor{tabsecond}1.43     \\ 
Ours                                                                  & \multicolumn{2}{c}{3DGS}                         & \cellcolor{tabfirst}1.09 & \cellcolor{tabfirst}3.26  & \cellcolor{tabfirst}2.65  & 0.95 & \cellcolor{tabfirst}0.15 & \cellcolor{tabfirst}0.35 & \cellcolor{tabfirst}0.44 & \cellcolor{tabsecond}0.39 & \cellcolor{tabfirst}2.47 & \cellcolor{tabsecond}1.19 & \cellcolor{tabfirst}0.95 & \cellcolor{tabfirst}1.40     \\ \bottomrule
\end{tabular}}}

  \caption{\textbf{Tracking performance comparison on KITTI dataset~\cite{kitti}.} The evaluation is conducted on the LiDAR dataset with motion compensated point cloud. We use ATE RMSE (m) as our metric. ``Avg." denotes the average results of all sequences. Note that ours is the only 3DGS-based method able to run successfully on all sequences, so we present only our method as a representative of GS-based approaches.}

  \label{tab:track3}
\end{table*}

\begin{table*}[]
\centering
\scalebox{0.95}{
\setlength{\tabcolsep}{1.5mm}{
\begin{tabular}{lcccccccc}
\hline
Methods    & Track/Iter. & Map/Iter. & Track/Frame(s) $\downarrow$ & Map/Frame(s) $\downarrow$ & Render FPS & Decoder Param. & Memory   & Peak GPU \\ \toprule
\multicolumn{9}{l}{\cellcolor[HTML]{EEEEEE}{\textit{NeRF-based Methods}}} \\
NICE-SLAM~\cite{niceslam}  & 6.98ms      & 28.88ms   & 68.54ms                     & 1.23s                     & 0.30       & 0.06M          & 48.48MB  & 12.0GB   \\
Vox-Fusion~\cite{voxfusion} & 7.02ms      & 23.34ms   & 350.9ms                       & \cellcolor{tabthird}0.47s                     & 1.31       & 0.06M          & 43.48MB  & 17.6GB   \\
Point-SLAM~\cite{pointslam} & \cellcolor{tabsecond}6.85ms      & 19.98ms   & \cellcolor{tabsecond}25.72ms                     & 10.47s                    & 1.33       & 0.127M         & 55.42MB  & \cellcolor{tabthird}8.5GB    \\
ESLAM~\cite{eslam}      & \cellcolor{tabsecond}6.85ms      & 19.98ms   & \cellcolor{tabthird}54.80ms                     & \cellcolor{tabsecond}0.29s                     & 2.82       & \cellcolor{tabthird}0.003M         & \cellcolor{tabsecond}27.12MB  & 17.3GB   \\
Co-SLAM~\cite{coslam}    & \cellcolor{tabfirst}6.38ms      & \cellcolor{tabsecond}14.25ms   & 63.93ms                     & \cellcolor{tabfirst}0.15s                     & 3.68       & 0.013M         & \cellcolor{tabfirst}24.85MB  & 16.7GB   \\
Loopy-SLAM~\cite{loopyslam} & 55.53ms     & \cellcolor{tabfirst}10.24ms   & 1.11s                       & 4.08s                     & 2.12       & 0.127M         & 60.98MB  & 13.3GB    \\ \midrule
\multicolumn{9}{l}{\cellcolor[HTML]{EEEEEE}{\textit{3DGS-based Methods}}} \\
SplaTAM~\cite{splatam}    & 24.23ms     & 22.83ms   & 2.18s                       & 1.37s                     & \cellcolor{tabthird}175.64     & \cellcolor{tabfirst}0M             & 273.09MB & 18.5GB   \\
Loop-Splat~\cite{loopsplat} & 10.28ms     & \cellcolor{tabthird}17.53ms   & 1.03s                       & 1.05s                     & \cellcolor{tabsecond}315.48     & \cellcolor{tabfirst}0M             & 93.87MB  & \cellcolor{tabsecond}7.4GB    \\
Ours       &  -           &     26.85ms      & \cellcolor{tabfirst}10.75ms                       & 0.93s                     & \cellcolor{tabfirst}322.45     & \cellcolor{tabthird}0.003M         & \cellcolor{tabthird}70.81MB     & \cellcolor{tabsecond}7.8GB    \\ \hline
\end{tabular}}}

\caption{\textbf{Runtime and Memory Usage on Replica } \texttt{Room 0}. Per-frame runtime is calculated as the total optimization time divided by the sequence length. ``-" is because our method does not use iterative tracking, making it infeasible to calculate per-iteration time. The memory usage represents the total memory of the map representation. ``Decoder Param." denotes decoder parameters. Note that implicit field-based methods require additional space for their decoders.}

\label{tab:time}
\end{table*}
\begin{table*}[]
\centering
\scalebox{0.96}{
\setlength{\tabcolsep}{1.0mm}{
\begin{tabular}{l|cccccc|cccccc}
\hline
\multirow{3}{*}{Method} & \multicolumn{6}{c|}{Replica~\cite{replica}}                                                                                 & \multicolumn{6}{c}{KITTI~\cite{kitti}}                                                                                    \\ \cline{2-13} 
                        & \multicolumn{2}{c}{Accuracy}        & \multicolumn{4}{c|}{Real-time performance and Memory}                  & \multicolumn{2}{c}{Accuracy}        & \multicolumn{4}{c}{Real-time performance and Memory}                   \\ \cline{2-13} 
                        & PSNR $\uparrow$ & RMSE $\downarrow$ & Time $\downarrow$ & Render FPS$\uparrow$ & Memory$\downarrow$ & GPU    & PSNR $\uparrow$ & RMSE $\downarrow$ & Time $\downarrow$ & Render FPS$\uparrow$ & Memory$\downarrow$ & GPU    \\ \hline
(a)w/o Vox.             & 33.08           & 0.24              & 0.53H             & 317.43               & 94.45MB            & 10.9GB & 21.01                 & 1.15                  & 3.71H                  & 261.39                     &  371.49MB             & 30.9GB \\
(b)w/o Prog.            & 31.49           & 0.27              & 0.51H             & 321.45               & 73.98MB            & 18.8GB & -                & -                 & -                  & -                     & -                    & -  \\
(c)only 2Dtrack         & 30.45           & 0.28              & 0.50H             & 322.45               & 70.81MB            & 7.6GB  & 17.92           & 5.96              & 3.02H             & 301.14              & 264.59MB                     & 19.9GB \\
(c)only 3Dtrack         & 30.21           & 0.41              & \textbf{0.41H}             & 322.45               & 70.81MB            & \textbf{7.3GB}  & 20.39                &     1.15              &  \textbf{2.61H}                 &    301.14                  &    264.59MB & \textbf{18.5GB}  \\
(d)w/o LC               & 32.87           & 0.23              & 0.49H             & 322.45               & 70.81MB            & 7.7GB  & 18.13                &     3.15              &  3.09H                 &    301.14                  &    264.59MB                & 20.2GB  \\
Full model        & \textbf{33.10}           & \textbf{0.21}              & 0.51H             & \textbf{322.45}               & \textbf{70.81MB}            & 7.8GB  & \textbf{21.37}                & \textbf{1.09}                  &  3.17H                 & \textbf{301.14}                     & \textbf{264.59MB}                   &  20.4GB  \\ \hline
\end{tabular}}}

\caption{Ablation study on the Replica~\cite{replica} and KITTI~\cite{kitti} dataset. ``H" denotes hours. ``-" denotes fail. The full model demonstrates superior pose estimation accuracy while maintaining faster training/rendering speed and lower memory consumption.}

\label{tab:ablation}
\end{table*}
\section{Experiments}

We validate that our method outperforms existing implicit representation-based methods in scene reconstruction, pose estimation, and real-time performance.

\noindent\textbf{Datasets.} We evaluate VPGS-SLAM on a variety of scenes from different datasets:
\begin{itemize}
    \item Replica Dataset~\cite{replica}. 8 small room scenes (\textbf{approximately} $6.5m\times4.2m\times2.7m $ ). We use this dataset to evaluate the reconstruction and localization accuracy in small-scale environments.
    \item ScanNet dataset~\cite{scannet}. Real-world scenes with long sequences (more than 5000 images) and large-scale indoor scenarios (\textbf{approximately} $7.5m\times6.6m\times3.5m $). We use this dataset for large-scale real-world indoor environments.
    \item KITTI dataset~\cite{kitti} and VKITTI 2 dataset~\cite{vkitti2}. Urban scenes dataset with long sequences (\textbf{approximately} $500m\times400m$). We use these datasets to validate the effectiveness of our method in virtual and real-world city-scale scenes.
    \item Newer College datasets~\cite{newer}. This dataset provides diverse trajectories and complex scenes, making it well-suited for assessing large-scale and high-fidelity reconstruction performance.
    \end{itemize}
    We use these datasets to validate the effectiveness of our method in indoor and real-world outdoor city-scale scenes.

\noindent\textbf{Implementation Details} 
We run our system on a desktop PC with NVIDIA RTX 3090 GPU. We set the voxelized corresponding parameter $k=10$. All the MLPs employed in our approach are 2-layer MLPs with ReLU activation; the dimensions of the hidden units are all 32. We set the submap initialization parameters with a distance threshold $d=0.5m$ in indoor scenes and $d=10m$ for outdoor scenes, and a rotation threshold $\omega =50$ degree. The two loss weight $\lambda_{SSIM}$ and $\lambda_{vol}$ are set to 0.2 and 0.001 in our experiments. We use DepthLab~\cite{depthlab} for dense depth image in outdoor scene.

\noindent\textbf{Scene Reconstruction} To evaluate the performance of scene reconstruction, we utilized three datasets: Replica for small room scenes, KITTI and vKITTI for large-scale urban scenes, and Newer College for campus environments.
We report the scene reconstruction results on Replica~\cite{replica} in Tab.~\ref{tab:render}. The best results are highlighted as \colorbox{tabfirst}{first}, \colorbox{tabsecond}{second}, and \colorbox{tabthird}{third}. Our method outperforms all 3DGS-based methods with superior rendering performance, demonstrating its effectiveness in reconstructing small room scenes. We also visualize the rendering performance in the Replica scene in Fig.~\ref{fig:replica1}. To evaluate the effectiveness of our method in large-scale urban scene reconstruction, we test it on the KITTI~\cite{kitti} and VKITTI2~\cite{vkitti2} dataset, as shown in Tab.~\ref{tab:kittirender}. In Fig.~\ref{fig:kitti1}, Fig.~\ref{fig:kitti2}, and Fig.~\ref{fig:kitti3}, we visualize the rendered results in different KITTI sequence and compare them with SplaTAM~\cite{splatam} and MonoGS~\cite{monogs}, demonstrating our clearly superior performance on scene reconstruction and rendering. We also visualize the mesh reconstruction results on the Newer College dataset in Fig. 1. The left image shows the reconstruction of the entire sequence, and the right image shows a zoomed-in view of a local area. As can be seen, in addition to the rendering performance, our mesh reconstruction also achieves excellent results.

\noindent\textbf{Camera Tracking}
We present our camera tracking performance on Replica~\cite{replica}, ScanNet~\cite{scannet}, and KITTI\cite{kitti} datasets in Tab.~\ref{tab:track1},~\ref{tab:track2},~\ref{tab:track3} and Fig.~\ref{fig:track1}. In the synthetic dataset Replica, our method surpasses existing state-of-the-art approaches, attributable to our 2D-3D fusion tracking method, which effectively integrates 3D Gaussian and 2D information. Our method achieves a 30$\%$ accuracy improvement on Replica dataset over current NeRF-based and 3DGS-based approaches. On the real-world indoor dataset ScanNet, our method also performs better than current NeRF-based and 3DGS-based approaches. In challenging sequences such as ScanNet00 and ScanNet69, where pose error accumulates continuously and loop closures occur, our method accurately identifies loop closure points and effectively mitigates accumulated pose errors. In large-scale outdoor datasets KITTI~\cite{kitti}, where most GS-based and NeRF-based methods fail, we compare our approach with pose estimation methods for large scenes, demonstrating a 10$\%$ accuracy improvement over NeRF-LOAM.

\noindent\textbf{Memory and Time analysis}
In Tab.~\ref{tab:time}, we present the runtime, memory usage, and Peak GPU memory of our method and other SOTA methods. In contrast to GS-based methods, our method employs a progressive map representation with multiple submaps, reducing excessive online GPU usage. This approach reduces GPU requirements by a factor of 2-3 compared to SplaTAM~\cite{splatam}, enabling our algorithm to be effectively utilized on edge computing platforms. Our voxelized scene representation method also significantly improves map representation efficiency, reducing the required memory six times compared to SplaTAM~\cite{splatam}.

\subsection{Ablation Study}

In this section, we validate the effectiveness of each module in our algorithm in both indoor and outdoor datasets, shown in Tab.~\ref{tab:ablation}. First, we explore the impact of voxelized scene representation and find that this module significantly enhances mapping efficiency in dense scenes. Next, we compare the effectiveness of progressive mapping with multiple submaps, which considerably reduces GPU computation requirements while improving tracking accuracy. This multi-map representation confines pose errors within each submap, effectively mitigating cumulative drift. We then assess the effectiveness of 2D-3D fusion tracking. Our findings indicate that 2D photometric loss performs better in indoor and simulated environments, whereas 3D geometric cues are more beneficial for outdoor scene tracking. Compared with 3D gaussian-based tracking, our tracking method can achieve more accurate matching with 2D-3D information fusion. Finally, we validate the loop closure module's effectiveness, showing that it significantly reduces cumulative pose errors.

\section{Conclusion}

In this paper, we propose a novel large-scale 3DGS SLAM framework, VPGS-SLAM, which achieves accurate scene reconstruction and pose estimation in both small and large-scale scenes. Our voxelized progressive mapping method achieves compact and accurate scene representation. The novel 2D-3D fusion camera tracking fully leverages the 3D Gaussian attributes with photometric information to efficiently match the 3D points and achieve accurate camera tracking. The 3D Gaussian loop closure enables global map consistency across multiple submaps and eliminates the accumulative pose error. The extensive experiments demonstrate the effectiveness and accuracy of our system in scene reconstruction, view synthesis, and pose estimation in various scenes.

\bibliographystyle{unsrt}  
\bibliography{neurips_2025}

  \begin{IEEEbiography}[{\includegraphics[width=1in,height=1.25in,clip,keepaspectratio]{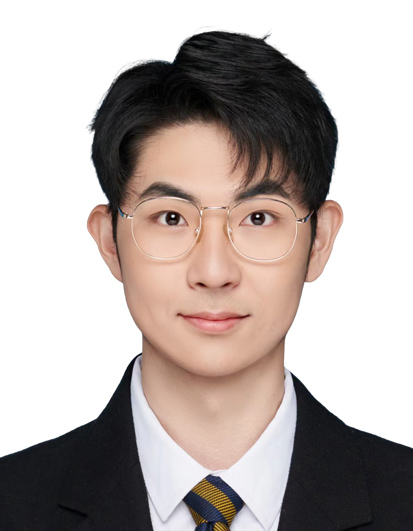}}]
 {Tianchen Deng} received the B.Eng. degree in control science and engineering from the Harbin Institute of Technology,
Harbin, China, in 2021. He is currently pursuing the
Joint Ph.D. degree in control science and engineering with
Shanghai Jiao Tong University and Nanyang Technological University.
His main research interests include 
visual SLAM, 3D Reconstruction, world model, and Embodied AI.
  \end{IEEEbiography}

  \begin{IEEEbiography}[{\includegraphics[width=1in,height=1.25in,clip,keepaspectratio]{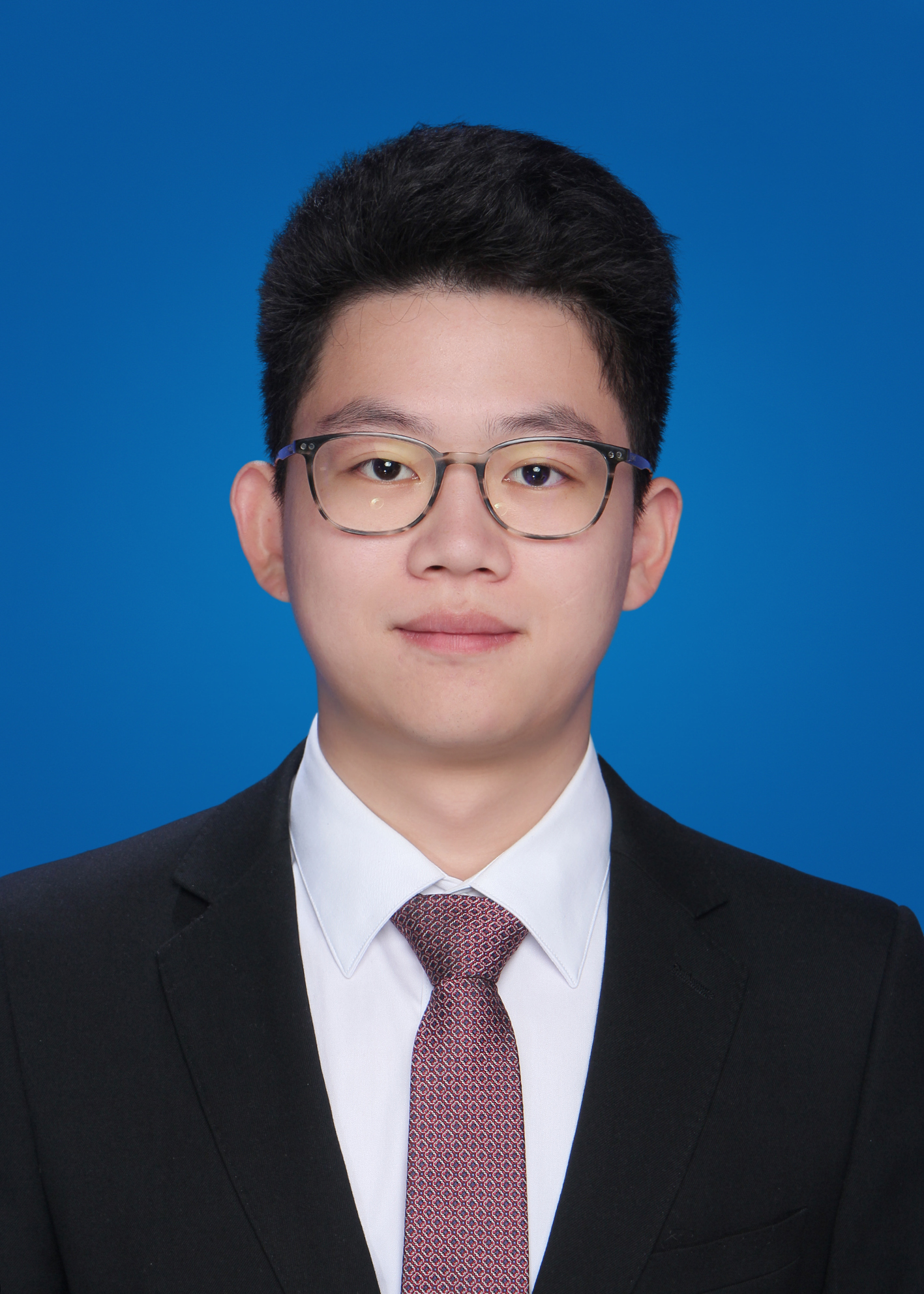}}]
 {Wenhua Wu} received the B.S. degree in Department of Automation, Shanghai Jiao Tong University, Shanghai, China, in 2023. He is currently pursuing the Ph.D. degree in Computer Science and Technology with Shanghai Jiao Tong University. His current research interests include robot learning and computer vision.
  \end{IEEEbiography}

   \begin{IEEEbiography}[{\includegraphics[width=1in,height=1.25in,clip,keepaspectratio]{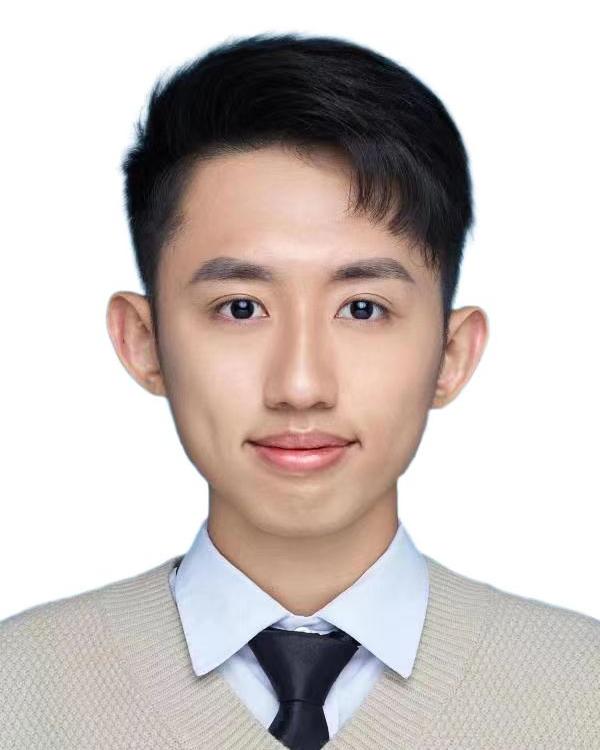}}]
 {Junjie He} received the B.E. degree in Automation from Guangdong University of Technology, in 2022, and the M.E. degree in artificial intelligence from Xi'an Jiaotong University, in 2025. From 2025, He works as a Research Assistant at the Thrust of Robotics and Autonomous Systems, The Hong Kong University of Science and Technology (Guangzhou). His research interests include 3D computer vision.
 \end{IEEEbiography}

   \begin{IEEEbiography}[{\includegraphics[width=1in,height=1.25in,clip,keepaspectratio]{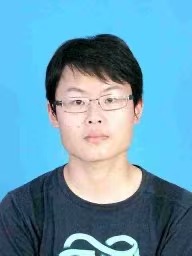}}]
 {Yue Pan} is a Ph.D. student at the Photogrammetry \& Robotics Lab at the University of Bonn, Germany. He obtained his B.Sc. degree in Geomatics Engineering from Wuhan University, China in 2019 and received his MSc degree in Geomatics Engineering from ETH Zurich, Switzerland in 2022. His research focuses on SLAM, 3D reconstruction, and navigation.
 \end{IEEEbiography}

   \begin{IEEEbiography}[{\includegraphics[width=1in,height=1.25in,clip,keepaspectratio]{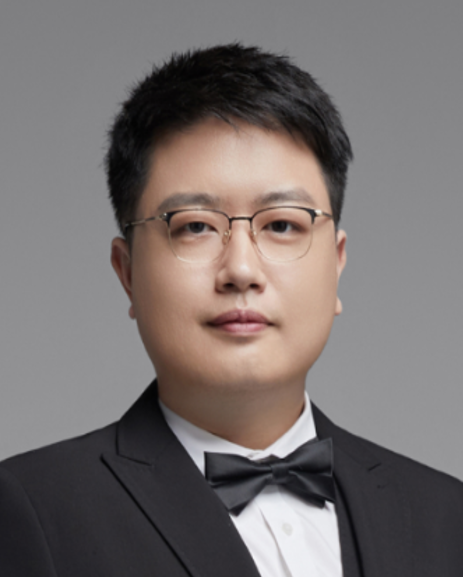}}]
 {Shenghai Yuan} is a senior research fellow at the Centre for Advanced Robotics Technology Innovation (CARTIN), Nanyang Technological University, Singapore. He received his B.S. and Ph.D. degrees in Electrical and Electronic Engineering in 2013 and 2019, respectively. His research focuses on robotics perception and navigation. Currently, he serves as an associate editor for the Unmanned Systems Journal and as a guest editor of the Electronics Special Issue on Advanced Technologies of Navigation for Intelligent Vehicles. He received the Outstanding Reviewer Award at ICRA 2024.

 \end{IEEEbiography}

     \begin{IEEEbiography}[{\includegraphics[width=1in,height=1.25in,clip,keepaspectratio]{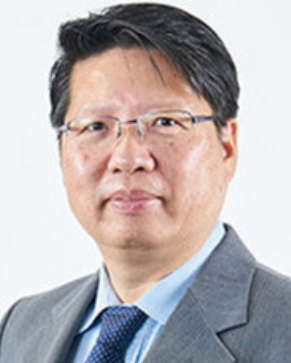}}]
 {Danwei Wang} (Life Fellow, IEEE) received the
B.E. degree from the South China University of
Technology, China, in 1982, and the M.S.E. and
Ph.D. degrees from the University of Michigan, Ann
Arbor, MI, USA, in 1984 and 1989, respectively. He is a fellow of the Academy of Engineering Singapore. He was a
recipient of the Alexander von Humboldt Fellowship, Germany. He served as
the general chairperson, the technical chairperson, and various positions for
several international conferences. He was an invited guest editor of various
international journals. He is a Distinguished Lecturer of the IEEE Robotics
and Automation Society.

 \end{IEEEbiography}
 
    \begin{IEEEbiography}[{\includegraphics[width=1in,height=1.25in,clip,keepaspectratio]{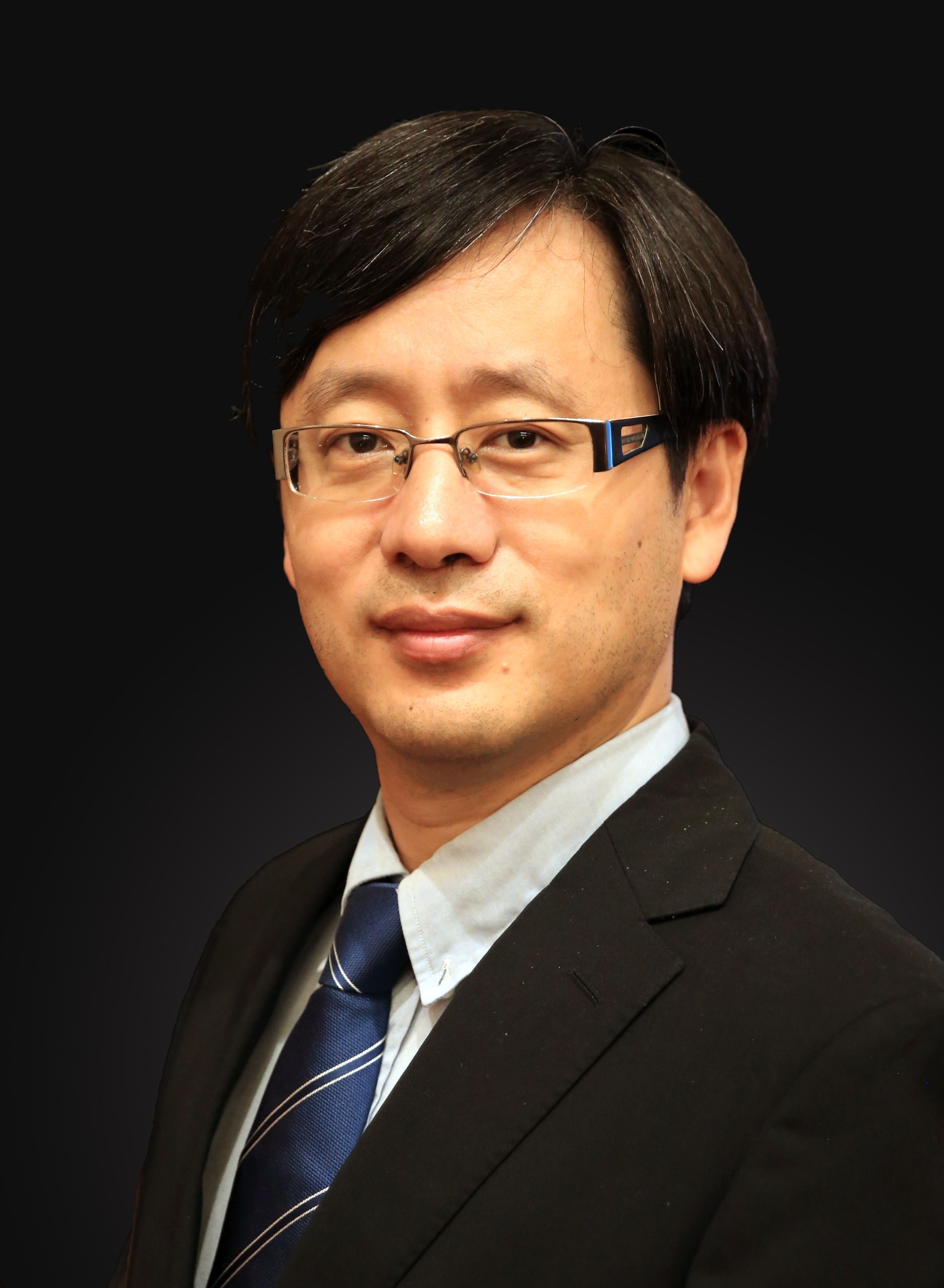}}]
 {Hesheng Wang}(Senior Member, IEEE) received
the B.Eng. degree in electrical engineering
from the Harbin Institute of Technology in
2002 and the M.Phil. and Ph.D. degrees in
automation and computer-aided engineering from
The Chinese University of Hong Kong, China, in
2004 and 2007, respectively.
He is currently a Professor with the Department of Automation, Shanghai Jiao Tong
University. He is an Associate Editor of Assembly
Automation and the International Journal of Humanoid Robotics and an Senior Editor of IEEE/ASME TRANSACTIONS ON MECHATRONICS. He served as an Associate Editor for IEEE TRANSACTIONS ON ROBOTICS
from 2015 to 2019. He was the General Chair of IEEE ROBIO 2022 and IEEE RCAR 2016, and the Program Chair of the IEEE
ROBIO 2014 and IEEE/ASME AIM 2019. He will be the General Chair of IEEE/RSJ IROS 2025.

 \end{IEEEbiography}
\end{document}